%% file: main.tex
\documentclass[runningheads]{eccv2024_stylekit/llncs}

\usepackage{eccv2024_stylekit/eccv}

\usepackage{eccv2024_stylekit/eccvabbrv}

\usepackage{graphicx}
\usepackage{booktabs}

\usepackage{amsmath}
\usepackage{amssymb}
\usepackage{multirow}
\usepackage{pifont}
\usepackage{soul}
\usepackage{lipsum} 
\usepackage{bm}
\usepackage{xcolor}
\usepackage{colortbl}
\usepackage{algorithm}
\usepackage{algpseudocode}
\usepackage{marvosym}
\usepackage[accsupp]{axessibility}  %
\usepackage{enumitem}

\usepackage[hyperfootnotes=false]{hyperref}

\usepackage{orcidlink}

\usepackage[capitalize]{cleveref}
\crefname{section}{Sec.}{Secs.}
\Crefname{section}{Section}{Sections}
\Crefname{table}{Table}{Tables}
\crefname{table}{Tab.}{Tabs.}

\newcommand{\supp}[1]{#1}
\newcommand{\new}[1]{#1}

\newcommand{\kk}[1]{}
\newcommand{\da}[1]{}
\newcommand{\ah}[1]{}
\newcommand{\ya}[1]{}
\newcommand{\ak}[1]{}
\newcommand{\fp}[1]{}

\renewcommand{\paragraph}[1]{\noindent\textbf{#1}}

\definecolor{rowx}{RGB}{242, 243, 244}
\definecolor{row}{RGB}{235, 245, 251}
\definecolor{suppcolor}{RGB}{0,0,255}
\newcommand{\pr}[1]{{\textcolor{red}{#1}}}

\newcommand{\fref}[1]{Fig.~\ref{#1}}
\newcommand{\tref}[1]{Table~\ref{#1}}
\newcommand{\sref}[1]{Sec.~\ref{#1}}

\newcommand{\tablestyle}[2]{\setlength{\tabcolsep}{#1}\renewcommand{\arraystretch}{#2}\centering\footnotesize}
\DeclareCaptionLabelFormat{andfigure}{\tablename~\thetable~\&~#1~#2}

\newcommand\blfootnote[1]{%
  \begingroup
  \renewcommand\thefootnote{}\footnote{#1}%
  \addtocounter{footnote}{-1}%
  \endgroup
}

\begin{document}

\title{Object-Centric Diffusion for Efficient \\Video Editing} 

\titlerunning{OCD: Object-Centric Diffusion}

\author{
Kumara Kahatapitiya$^*$\and
Adil Karjauv\and
Davide Abati\and 
Fatih Porikli\and\\
Yuki M. Asano\and
Amirhossein Habibian
}

\authorrunning{Kahatapitiya et al.}

\institute{
Qualcomm AI Research$^{\dagger}$
\email{\{kkahatap,akarjauv,dabati,fporikli,asano,ahabibian\}@qti.qualcomm.com}
}
\input{macros}
\maketitle
\input{texts/0_abstract}
\blfootnote{$^*$Work completed during internship at Qualcomm Technologies, Inc}
\blfootnote{$^{\dagger}$Qualcomm AI Research is an initiative of Qualcomm Technologies,
Inc.}
\input{texts/1_introduction}
\input{texts/2_related_work}
\input{texts/3_analysis}
\input{texts/4_method}
\input{texts/5_experiments}
\input{texts/6_conclusion}
\bibliographystyle{eccv2024_stylekit/splncs04}
\bibliography{main.bbl}
\clearpage
\begin{center}
\large
\textbf{Object-Centric Diffusion for Efficient Video Editing: Supplementary Material}
\end{center}
\input{texts/supplementary/A_discussion}
\input{texts/supplementary/B_results}

\end{document}

%% file: macros.tex
\newcommand{\head}[1]{{\smallskip\noindent\textbf{#1}}}
\newcommand{\alert}[1]{{\color{red}{#1}}}
\newcommand{\sm}{\scriptsize}
\newcommand{\eq}[1]{(\ref{eq:#1})}

\newcommand{\Th}[1]{\textsc{#1}}
\newcommand{\mr}[2]{\multirow{#1}{*}{#2}}
\newcommand{\mc}[2]{\multicolumn{#1}{c}{#2}}
\newcommand{\tb}[1]{\textbf{#1}}
\newcommand{\ch}{\checkmark}

\newcommand{\red}[1]{{\color{red}{#1}}}
\newcommand{\blue}[1]{{\color{blue}{#1}}}
\newcommand{\green}[1]{\color{green}{#1}}
\newcommand{\gray}[1]{{\color{gray}{#1}}}

\newcommand{\citeme}[1]{\red{[XX]}}
\newcommand{\refme}[1]{\red{(XX)}}

\newcommand{\fig}[2][1]{\includegraphics[width=#1\linewidth]{fig/#2}}
\newcommand{\figh}[2][1]{\includegraphics[height=#1\linewidth]{fig/#2}}

\newcommand{\tran}{^\top}
\newcommand{\mtran}{^{-\top}}
\newcommand{\zcol}{\mathbf{0}}
\newcommand{\zrow}{\zcol\tran}

\newcommand{\ind}{\mathbbm{1}}
\newcommand{\expect}{\mathbb{E}}
\newcommand{\nat}{\mathbb{N}}
\newcommand{\zahl}{\mathbb{Z}}
\newcommand{\real}{\mathbb{R}}
\newcommand{\proj}{\mathbb{P}}
\newcommand{\prob}{\mathbf{Pr}}
\newcommand{\normal}{\mathcal{N}}

\newcommand{\mif}{\textrm{if}\ }
\newcommand{\other}{\textrm{otherwise}}
\newcommand{\minimize}{\textrm{minimize}\ }
\newcommand{\maximize}{\textrm{maximize}\ }

\newcommand{\id}{\operatorname{id}}
\newcommand{\const}{\operatorname{const}}
\newcommand{\sgn}{\operatorname{sgn}}
\newcommand{\var}{\operatorname{Var}}
\newcommand{\mean}{\operatorname{mean}}
\newcommand{\trace}{\operatorname{tr}}
\newcommand{\diag}{\operatorname{diag}}
\newcommand{\vect}{\operatorname{vec}}
\newcommand{\cov}{\operatorname{cov}}
\newcommand{\sign}{\operatorname{sign}}
\newcommand{\prj}{\operatorname{proj}}

\newcommand{\softmax}{\operatorname{softmax}}
\newcommand{\clip}{\operatorname{clip}}

\newcommand{\defn}{\mathrel{:=}}
\newcommand{\peq}{\mathrel{+\!=}}
\newcommand{\meq}{\mathrel{-\!=}}

\newcommand{\floor}[1]{\left\lfloor{#1}\right\rfloor}
\newcommand{\ceil}[1]{\left\lceil{#1}\right\rceil}
\newcommand{\inner}[1]{\left\langle{#1}\right\rangle}
\newcommand{\norm}[1]{\left\|{#1}\right\|}
\newcommand{\abs}[1]{\left|{#1}\right|}
\newcommand{\frob}[1]{\norm{#1}_F}
\newcommand{\card}[1]{\left|{#1}\right|\xspace}
\newcommand{\divg}[2]{{#1\ ||\ #2}}
\newcommand{\diff}{\mathrm{d}}
\newcommand{\der}[3][]{\frac{d^{#1}#2}{d#3^{#1}}}
\newcommand{\pder}[3][]{\frac{\partial^{#1}{#2}}{\partial{#3^{#1}}}}
\newcommand{\ipder}[3][]{\partial^{#1}{#2}/\partial{#3^{#1}}}
\newcommand{\dder}[3]{\frac{\partial^2{#1}}{\partial{#2}\partial{#3}}}

\newcommand{\wb}[1]{\overline{#1}}
\newcommand{\wt}[1]{\widetilde{#1}}

\def\xssp{\hspace{1pt}}
\def\ssp{\hspace{3pt}}
\def\msp{\hspace{5pt}}
\def\lsp{\hspace{12pt}}

\newcommand{\cA}{\mathcal{A}}
\newcommand{\cB}{\mathcal{B}}
\newcommand{\cC}{\mathcal{C}}
\newcommand{\cD}{\mathcal{D}}
\newcommand{\cE}{\mathcal{E}}
\newcommand{\cF}{\mathcal{F}}
\newcommand{\cG}{\mathcal{G}}
\newcommand{\cH}{\mathcal{H}}
\newcommand{\cI}{\mathcal{I}}
\newcommand{\cJ}{\mathcal{J}}
\newcommand{\cK}{\mathcal{K}}
\newcommand{\cL}{\mathcal{L}}
\newcommand{\cM}{\mathcal{M}}
\newcommand{\cN}{\mathcal{N}}
\newcommand{\cO}{\mathcal{O}}
\newcommand{\cP}{\mathcal{P}}
\newcommand{\cQ}{\mathcal{Q}}
\newcommand{\cR}{\mathcal{R}}
\newcommand{\cS}{\mathcal{S}}
\newcommand{\cT}{\mathcal{T}}
\newcommand{\cU}{\mathcal{U}}
\newcommand{\cV}{\mathcal{V}}
\newcommand{\cW}{\mathcal{W}}
\newcommand{\cX}{\mathcal{X}}
\newcommand{\cY}{\mathcal{Y}}
\newcommand{\cZ}{\mathcal{Z}}

\newcommand{\vA}{\mathbf{A}}
\newcommand{\vB}{\mathbf{B}}
\newcommand{\vC}{\mathbf{C}}
\newcommand{\vD}{\mathbf{D}}
\newcommand{\vE}{\mathbf{E}}
\newcommand{\vF}{\mathbf{F}}
\newcommand{\vG}{\mathbf{G}}
\newcommand{\vH}{\mathbf{H}}
\newcommand{\vI}{\mathbf{I}}
\newcommand{\vJ}{\mathbf{J}}
\newcommand{\vK}{\mathbf{K}}
\newcommand{\vL}{\mathbf{L}}
\newcommand{\vM}{\mathbf{M}}
\newcommand{\vN}{\mathbf{N}}
\newcommand{\vO}{\mathbf{O}}
\newcommand{\vP}{\mathbf{P}}
\newcommand{\vQ}{\mathbf{Q}}
\newcommand{\vR}{\mathbf{R}}
\newcommand{\vS}{\mathbf{S}}
\newcommand{\vT}{\mathbf{T}}
\newcommand{\vU}{\mathbf{U}}
\newcommand{\vV}{\mathbf{V}}
\newcommand{\vW}{\mathbf{W}}
\newcommand{\vX}{\mathbf{X}}
\newcommand{\vY}{\mathbf{Y}}
\newcommand{\vZ}{\mathbf{Z}}

\newcommand{\va}{\mathbf{a}}
\newcommand{\vb}{\mathbf{b}}
\newcommand{\vc}{\mathbf{c}}
\newcommand{\vd}{\mathbf{d}}
\newcommand{\ve}{\mathbf{e}}
\newcommand{\vf}{\mathbf{f}}
\newcommand{\vg}{\mathbf{g}}
\newcommand{\vh}{\mathbf{h}}
\newcommand{\vi}{\mathbf{i}}
\newcommand{\vj}{\mathbf{j}}
\newcommand{\vk}{\mathbf{k}}
\newcommand{\vl}{\mathbf{l}}
\newcommand{\vm}{\mathbf{m}}
\newcommand{\vn}{\mathbf{n}}
\newcommand{\vo}{\mathbf{o}}
\newcommand{\vp}{\mathbf{p}}
\newcommand{\vq}{\mathbf{q}}
\newcommand{\vr}{\mathbf{r}}
\newcommand{\Vs}{\mathbf{s}}
\newcommand{\vt}{\mathbf{t}}
\newcommand{\vu}{\mathbf{u}}
\newcommand{\vv}{\mathbf{v}}
\newcommand{\vw}{\mathbf{w}}
\newcommand{\vx}{\mathbf{x}}
\newcommand{\vy}{\mathbf{y}}
\newcommand{\vz}{\mathbf{z}}

\newcommand{\vone}{\mathbf{1}}
\newcommand{\vzero}{\mathbf{0}}

\newcommand{\valpha}{{\boldsymbol{\alpha}}}
\newcommand{\vbeta}{{\boldsymbol{\beta}}}
\newcommand{\vgamma}{{\boldsymbol{\gamma}}}
\newcommand{\vdelta}{{\boldsymbol{\delta}}}
\newcommand{\vepsilon}{{\boldsymbol{\epsilon}}}
\newcommand{\vzeta}{{\boldsymbol{\zeta}}}
\newcommand{\veta}{{\boldsymbol{\eta}}}
\newcommand{\vtheta}{{\boldsymbol{\theta}}}
\newcommand{\viota}{{\boldsymbol{\iota}}}
\newcommand{\vkappa}{{\boldsymbol{\kappa}}}
\newcommand{\vlambda}{{\boldsymbol{\lambda}}}
\newcommand{\vmu}{{\boldsymbol{\mu}}}
\newcommand{\vnu}{{\boldsymbol{\nu}}}
\newcommand{\vxi}{{\boldsymbol{\xi}}}
\newcommand{\vomikron}{{\boldsymbol{\omikron}}}
\newcommand{\vpi}{{\boldsymbol{\pi}}}
\newcommand{\vrho}{{\boldsymbol{\rho}}}
\newcommand{\vsigma}{{\boldsymbol{\sigma}}}
\newcommand{\vtau}{{\boldsymbol{\tau}}}
\newcommand{\vupsilon}{{\boldsymbol{\upsilon}}}
\newcommand{\vphi}{{\boldsymbol{\phi}}}
\newcommand{\vchi}{{\boldsymbol{\chi}}}
\newcommand{\vpsi}{{\boldsymbol{\psi}}}
\newcommand{\vomega}{{\boldsymbol{\omega}}}

\newcommand{\rLambda}{\mathrm{\Lambda}}
\newcommand{\rSigma}{\mathrm{\Sigma}}

\newcommand{\vLambda}{\bm{\rLambda}}
\newcommand{\vSigma}{\bm{\rSigma}}

\makeatletter
\newcommand{\vast}[1]{\bBigg@{#1}}
\makeatother

\makeatletter
\newcommand*\bdot{\mathpalette\bdot@{.7}}
\newcommand*\bdot@[2]{\mathbin{\vcenter{\hbox{\scalebox{#2}{$\m@th#1\bullet$}}}}}
\makeatother

\makeatletter
\DeclareRobustCommand\onedot{\futurelet\@let@token\@onedot}
\def\@onedot{\ifx\@let@token.\else.\null\fi\xspace}

\def\eg{\emph{e.g}\onedot} \def\Eg{\emph{E.g}\onedot}
\def\ie{\emph{i.e}\onedot} \def\Ie{\emph{I.e}\onedot}
\def\cf{\emph{cf}\onedot} \def\Cf{\emph{Cf}\onedot}
\def\etc{\emph{etc}\onedot} \def\vs{\emph{vs}\onedot}
\def\wrt{w.r.t\onedot} \def\dof{d.o.f\onedot} \def\aka{a.k.a\onedot}
\def\etal{\emph{et al}\onedot}
\makeatother

\newcommand{\cmark}{\ding{51}}%
\newcommand{\xmark}{\ding{55}}%

\newcommand{\z}{\mathbf{z}}
\newcommand{\x}{\mathbf{x}}

%% file: texts/0_abstract.tex
\begin{abstract}
Diffusion-based video editing have reached impressive quality and can transform either the global style, local structure, and attributes of given video inputs, following textual edit prompts.
However, such solutions typically incur heavy memory and computational costs to generate temporally-coherent frames, either in the form of diffusion inversion and/or cross-frame attention.
In this paper, we conduct an analysis of such inefficiencies, and suggest simple yet effective modifications that allow significant speed-ups whilst maintaining quality.
Moreover, we introduce Object-Centric Diffusion, to \new{fix generation artifacts} and further reduce latency by allocating more computations towards foreground edited regions, arguably more important for perceptual quality.
We achieve this by two novel proposals: i) Object-Centric Sampling, decoupling the diffusion steps spent on salient or background regions and spending most on the former, and ii) Object-Centric Token Merging, which reduces cost of cross-frame attention by fusing redundant tokens in unimportant background regions.
Both techniques are readily applicable to a given video editing model \textit{without} retraining, and can drastically reduce its memory and computational cost. We evaluate our proposals on inversion-based and control-signal-based editing pipelines, and show a latency reduction up to 10$\times$ for a comparable synthesis quality.  Project page: \href{https://qualcomm-ai-research.github.io/object-centric-diffusion}{\texttt{qualcomm-ai-research.github.io/object-centric-diffusion}}.
\keywords{Video Editing \and Efficiency \and Object-Centric \and Diffusion}
\end{abstract}

%% file: texts/1_introduction.tex
\section{Introduction}
\input{figures/teaser}
Diffusion models~\cite{ddim, ddpm} stand as the fundamental pillar of contemporary generative AI approaches~\cite{stablediffusion, ho2022cascaded, kim2023consistency}. Their success primarily stems from their unparalleled diversity and synthesis quality, surpassing the capabilities of earlier versions of generative models~\cite{vae, gan}. On top of that, recent methods such as Latent Diffusion Models (LDMs)~\cite{stablediffusion} exhibit scalability to high-resolution inputs and adaptability to different domains without requiring retraining or finetuning, especially when trained on large-scale datasets~\cite{laion}. 
These traits catalyzed impactful adaptations of pretrained LDMs across a spectrum of applications, including text-guided image generation~\cite{stablediffusion}, editing~\cite{avrahami2023blended}, inpainting~\cite{stablediffusion}, as well as video generation~\cite{align_your_latents} and editing~\cite{tuneavideo, fatezero, controlvideo}.

Nevertheless, diffusion models come with various trade-offs. One immediate drawback is their inefficient sampling process, which involves iterating a denoising neural network across numerous diffusion steps. Despite the availability of techniques like step distillation~\cite{stepdist, progdist} and accelerated samplers~\cite{dpm, dpm++, ddim} that expedite image synthesis, efficient sampling solutions for video generation are still lacking. Besides, special attention must be paid to maintaining temporal coherency among frames when creating or modifying a video, in order to avoid flickering artifacts or lack of correlation between frames. To this aim, approaches like diffusion inversion~\cite{ddim,fatezero} and cross-frame self-attention~\cite{fatezero, controlvideo} have been introduced, but all  at the cost of further increasing the computational load.

This paper centers on video editing models and presents novel solutions to enhance their efficiency. We first examine the current video editing frameworks, identifying the key elements that increase their latency. These encompass memory overheads, such as those associated with attention-based guidance from diffusion inversion, as well as computational bottlenecks, including excessive cross-frame attention and an unnecessarily high number of sampling steps. We show that significant improvements can be achieved by adopting off-the-shelf optimizations namely efficient samplers and leveraging token reduction techniques in attention layers such as token merging (ToMe)~\cite{bolya2022token, bolya2023token}.

Additionally, we argue that video editing users may be particularly sensitive to the quality of edited foreground objects, as opposed to slight degradations in background regions. 
Consequently, we propose two new and efficient techniques that harness this object-centric aspect of editing applications. Our first solution, \textit{Object-Centric Sampling}, involves separating the diffusion process between edited objects and background regions. 
This strategy enables the model to focus most of its diffusion steps on the foreground areas, whilst generating unedited regions with considerably fewer steps, yet maintaining faithful reconstructions in all areas. With our second solution, \textit{Object-Centric ToMe}, we incentivise each token reduction layer to merge more tokens within less crucial background areas and exploit temporal redundancies that are abundant in videos.
The combination of such techniques, coined Object-Centric Diffusion (OCD), can seamlessly be applied to any video editing method without retraining or finetuning.

As we demonstrate in extensive experiments, the combined implementation of these object-centric solutions enhances the quality of video editing in both salient foreground object and background, while considerably reducing the generation cost. We accelerate both inversion-based and ControlNet-based video editing models, speeding them up by a factor of 10$\times$ and 6$\times$ respectively while also decreasing memory consumption up to 17$\times$ for a comparable quality (see~\cref{fig:teaser}). %

\noindent We summarize our contributions as follows:
\begin{itemize}[leftmargin=15pt,topsep=0pt,noitemsep]
\item We analyze the cost and inefficiencies of recent inversion-based video editing methods, and suggest simple ways to considerably speed them up.
\item We introduce Object-Centric Sampling, which separates diffusion sampling for the edited objects and background areas, limiting most denoising steps to the former to improve efficiency.
\item We introduce Object-Centric ToMe, which reduces number of cross-frame attention tokens by encouraging their fusion in background regions.
\item We showcase the effectiveness of OCD by optimizing two recent video editing models, obtaining very fast editing speed without compromising fidelity.%
\end{itemize}

%% file: figures/teaser.tex
\begin{figure}[t!]
\centering
\includegraphics[width=0.6\linewidth]{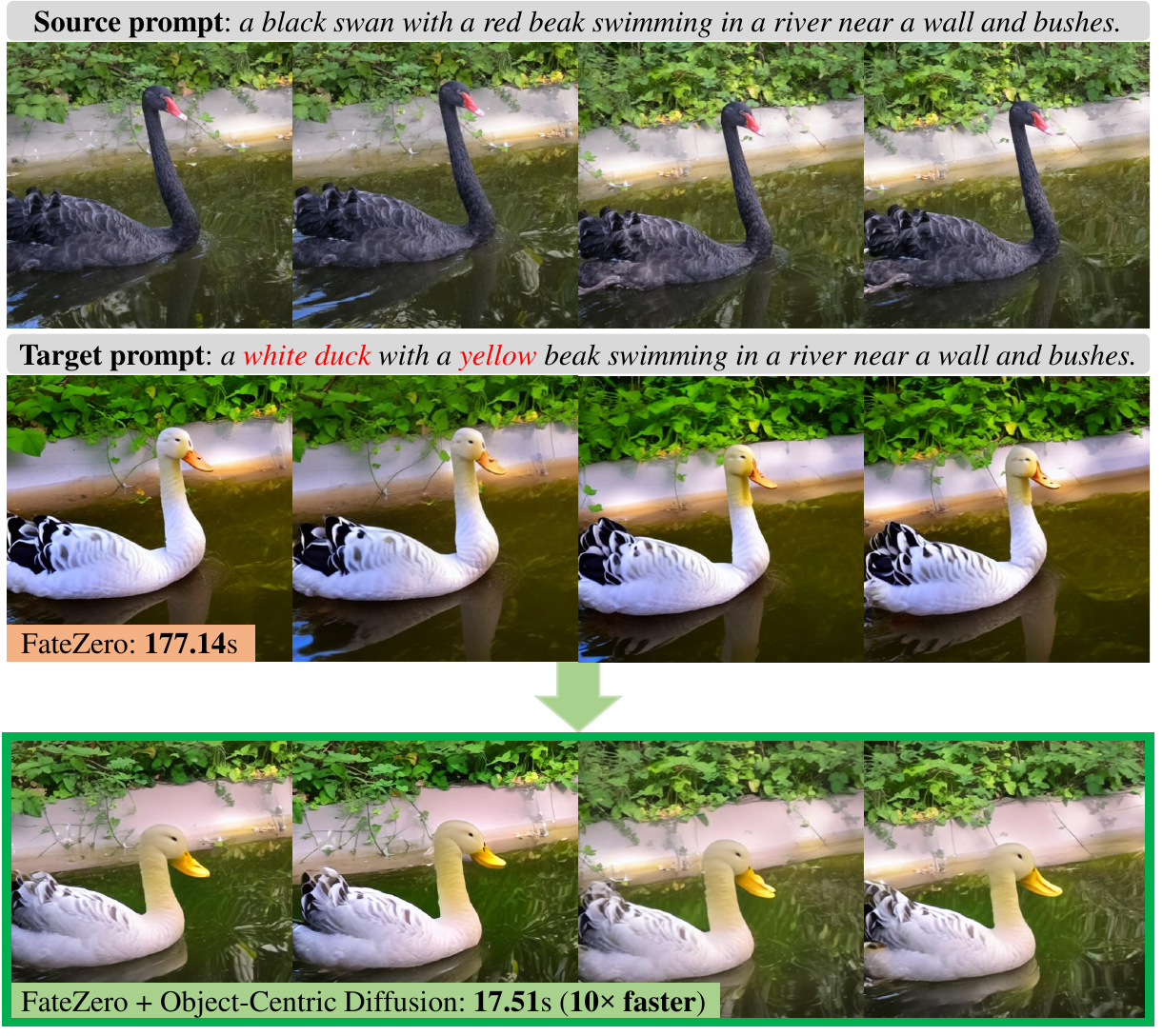}
\caption{
\textbf{OCD speeds up video editing.}
    We show exemplar editing results of FateZero~\cite{fatezero} with and without our OCD optimizations. When including our techniques, the editing is $10\times$ faster than the baseline with similar generation quality.}
\label{fig:teaser}
\end{figure}

%% file: texts/2_related_work.tex
\section{Related work}
\paragraph{Text-based Video Generation \& Editing} 
There has been a surge in text-to-video generation methods using diffusion models.
The early works inflate the image generation architectures by replacing most operations,~\ie convolutions and transformers, with their spatio-temporal counterparts~\cite{vdm, imagenvideo, hong2022cogvideo, makeavideo, magicvideo}. Despite their high temporal consistency, these substantial model modifications require extensive model training on large collection of captioned videos.

To avoid extensive model trainings, recent works adopt off-the-shelf image generation models for video editing in one-shot and zero-shot settings. One-shot methods rely on test-time model adaption on the editing sample, which is unfeasible for real-time applications~\cite{tuneavideo, esser2023structure}. Zero-shot methods integrate training-free techniques into the image generation models to ensure temporal consistency across frames~\cite{fatezero, controlvideo, text2video-zero, tokenflow, video-p2p} most commonly by: \emph{i)} providing strong structural conditioning from the input video~\cite{esser2023structure, controlvideo, control-a-video} inspired by ControlNet~\cite{controlnet}; \emph{ii)} injecting the activations and attention maps extracted by diffusion inversion~\cite{ddim, plugandplay} of the input video into the generation~\cite{fatezero, tokenflow}; \emph{iii)} replacing the self-attention operations in the architecture with temporal counterparts operating on neighboring frames~\cite{tuneavideo,fatezero, controlvideo}. Introducing cross-frame attention greatly increases the temporal consistency, especially when involving more and more frames in the operation~\cite{controlvideo}. Despite their effectiveness, all these solutions come with additional computational costs, which are currently underexplored in the literature.

\paragraph{Efficient Diffusion Models}
Due to their sequential sampling, diffusion models are computationally very expensive. %
Several studies analyze the denoising U-Net architecture to enhance its runtime efficiency, achieved either through model distillation~\cite{kim2023architectural} or by eliminating redundant modules~\cite{snapfusion}.
Other ways to obtain efficiency gains include enhanced noise schedulers~\cite{dpm, dpm++} and step distillation techniques~\cite{progdist, snapfusion, stepdist}.
Moreover, the work in~\cite{li2022efficient} enables efficiency in image editing applications by caching representations of regions left unedited, yet it cannot handle videos and it strictly requires human-in-the-loop to operate.
The most relevant to our work is the Token Merging technique presented in~\cite{bolya2022token} that demonstrated advantages for image generation~\cite{bolya2023token}. 
Although very proficient, we observe it does not readily work when deployed in video editing settings.
For this reason, we optimize this technique for the task at hand, such that it exploits redundancy of tokens across frames and directing its token fusing towards background regions.
\new{
Finally, Token Merging was applied to video editing use cases in VidToMe~\cite{li2023vidtome}, where its introduction is aimed at improving temporal consistency of generated frames.
Differently, we tailor it to improve efficiency without sacrificing fidelity in edited objects, resulting in much faster generations, as we shall demonstrate in our experimentation.
}

%% file: texts/3_analysis.tex
\section{Efficiency Bottlenecks} 
\label{sec:analysis}
To investigate the main latency bottlenecks in video editing pipelines, we run a benchmark analysis on a representative pipeline, derived from FateZero~\cite{fatezero}.
More specifically, we are interested in the role of components such as diffusion inversion and cross-frame attention, as these techniques are crucial for the task and ubiquitous in literature~\cite{tuneavideo,controlvideo,fatezero,tokenflow,li2023vidtome}.
To study their impact, we perform text-guided edits on 8-frame clips while varying (1) the number of diffusion steps (50$\rightarrow$20), and (2) the span of cross-frame attention, \ie self (1-frame), sparse (2-frames) or dense (8-frames).
In doing so, we probe the model's latency using DeepSpeed~\cite{deepspeed}.
Based on the latency measurements aggregated over relevant groups of operations as reported in~\fref{fig:probe_graph}, we make the following observations:

\paragraph{Observation 1.} As testified by~\cref{fig:probe_graph} (a), inversion-based pipelines are susceptible to memory operations. 
Not only the latency due to memory access dominates the overall latency in all the tested settings, but it also scales exponentially with the span (i.e., \#frames) of cross-frame attention.

\paragraph{Observation 2.} When ruling out memory and considering computations only, as in~\cref{fig:probe_graph} (b) and~\cref{fig:probe_graph} (c), we appreciate that attention-related operations are as expensive as (or often more) than all other network operations combined (\ie convolutions, normalizations, MLPs etc.).
This latter finding is consistent both in inversion and generation phases.

\paragraph{Observation 3.} More-powerful cross-frame attention operations (i.e., dense), that increase the temporal stability of generations~\cite{controlvideo}, come with a high latency cost, as observed in~\cref{fig:probe_graph} (b) and~\cref{fig:probe_graph} (c).
\input{figures/analysis}
\subsection{Off-the-shelf acceleration}
\label{sec:offtheshelf}
We explore whether the solutions available for image generation mitigate the key computational bottlenecks observed for video editing.
More specifically, we study whether token merging~\cite{bolya2022token}, as an effective strategy to improve the attention cost for image generation~\cite{bolya2023token}, reduces the high memory and computational cost of video editing. Moreover, given the linear relation between the number of diffusion steps and memory costs of inversion-based editing to store the attention maps and activations~\cite{plugandplay, fatezero,tokenflow, tuneavideo}, we explore how using faster noise schedulers improves the video editing efficiency. In what follows, we describe how the off-the-shelf accelerations are adopted for video editing:

\paragraph{Faster self-attention}
Aiming to speed-up attention operations, we utilize Token Merging (ToMe)~\cite{bolya2022token,bolya2023token} to merge redundant tokens.
More specifically, ToMe first splits the input self-attention tokens into source (\textit{src}) and destination (\textit{dst}) sets.
Then, the similarity between the source token $\x_i$ and the destination token $\x_j$ is computed based on normalized cosine similarity as follows:
\begin{equation}
\label{eq:similarity_no_saliency}
\text{Sim}(\x_i,\x_j)=\frac{1}{2}\bigg[\frac{\x_i \cdot \x_j}{\norm{\x_i}\norm{\x_j}} + 1\bigg].
\end{equation}
Finally, the top $r\%$ similarities are selected, and source tokens are piggybacked into the corresponding destination tokens by means of average pooling.
This process lightens successive operations as only destination and unmerged-source tokens (\textit{unm}) are forwarded, at the cost of a slight information loss.

We observed the potential of ToMe in speeding up video editing, as a simple reduction of $1/16$ in key-value tokens reduces the model latency by a factor of 4$\times$.
Yet, its naive application completely breaks generation quality, as testified in~\cref{fig:naive_tome} (d).
However, we could restore an outstanding generation quality by applying two simple yet crucial tricks.
Specifically, we \emph{i)} pair the token locations (both \textit{dst} and \textit{unm}), for the same network layer and diffusion timestep, between inversion and generation (see ToMe-paired~\cref{fig:naive_tome}(e)), and \emph{ii)} we resample the destination token locations for every frame (\cref{fig:naive_tome} (f)).
We refer the reader to the \supp{supplementary material} for more details about these ToMe optimizations.

\paragraph{Faster noise scheduler}
Although we observe that the commonly-used DDIM scheduler~\cite{ddim} suffers considerable quality degradation when reducing the number of diffusion steps (without any finetuning), 
we notice this is not the case with a more modern DPM++~\cite{dpm++}, that can operate with as few as 20 iterations.
As illustrated in the example reported in~\cref{fig:naive_tome} (c), this replacement does not affect the quality of the editing, yet it impacts the generation latency tremendously.
\\\\
We will refer to models optimized with these off-the-shelf accelerations as \emph{optimized}, and will serve as strong baselines to our proposed Object Centric Diffusion.
\new{Such a baseline is extremely fast, yet prone to sporadic generation artifacts (\cref{fig:naive_tome} (f), \cref{fig:ablations_3d.v.2d}), that we shall fix with the following object-centric solutions.}
\input{figures/naive_tome}

%% file: figures/analysis.tex
\begin{figure}[t!]
\centering
\includegraphics[width=0.6\linewidth]{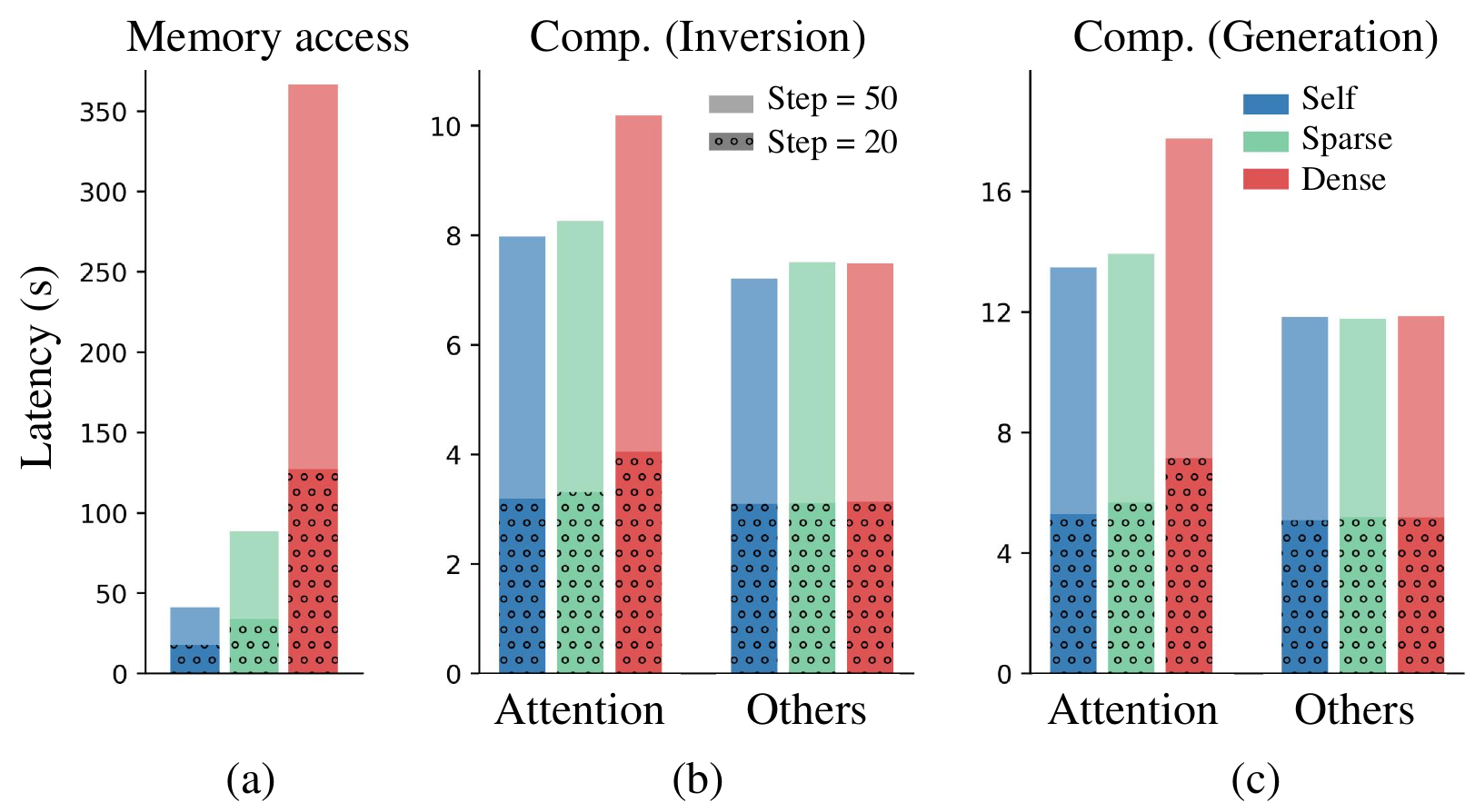}
\caption{
\textbf{Latency analysis} of video editing models. 
At various diffusion steps, latency is dominated by memory access operations. 
Among pure computations, attention alone is the bottleneck, especially when using dense cross-frame interactions.
As attention is the main responsible for most of the memory overhead, we hypothesize that reducing the number of its tokens have a significant impact on latency.
}
\label{fig:probe_graph}
\end{figure}

%% file: figures/naive_tome.tex
\begin{figure*}[tb]
\centering
\includegraphics[width=0.9\textwidth]{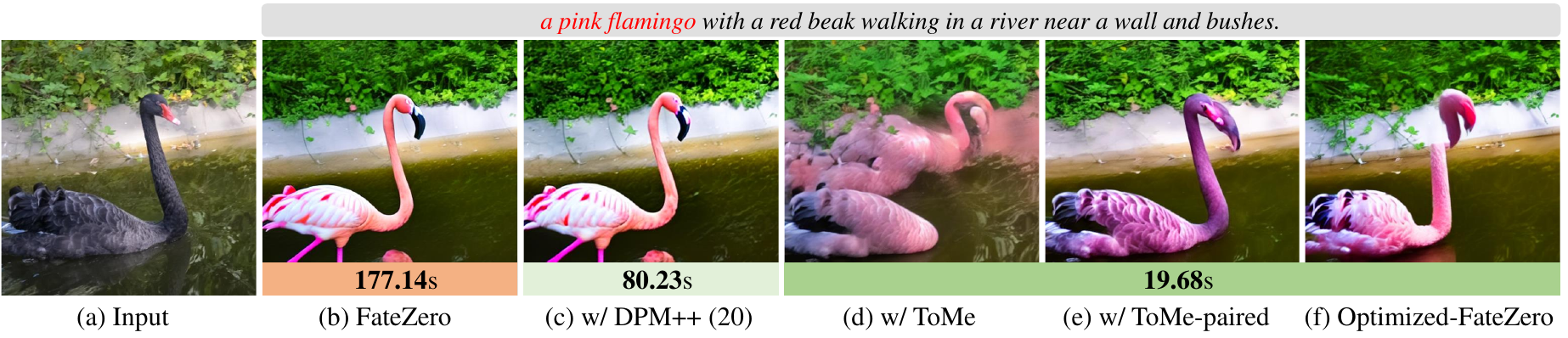}
\caption{
\textbf{Off-the-shelf accelerations:} 
First, we replace the (b) default sampler with (c) DPM++ \cite{dpm,dpm++}, allowing to reduce sampling steps from 50$\rightarrow$20 without a heavy degradation. 
Then, by applying ToMe, memory and computational overhead decreases, yet results degrade significantly (d). 
We therefore implement (e) pairing of ToMe indexes between inversion and generation and (f) per-frame resampling of destination tokens, regaining the quality.
Altogether, we coin the resulting model \textit{Optimized-FateZero}.
}
\label{fig:naive_tome}
\end{figure*}

%% file: texts/4_method.tex
\section{Object-Centric Diffusion}

Although off-the-shelf optimizations can notably improve latency, \new{they occasionally introduce some editing artifacts in foreground objects, that we solve by exploiting the object-centric nature of video editing tasks, also enabling further reductions in latency.}
Following our assumption regarding the significance of foreground objects for video editing applications, we propose two add-ons for diffusion models to channel their computational capacity towards foreground regions, and we describe them in \cref{method:oc_sampling} and in \cref{method:oc_merging}.
We will assume access to a foreground mask $m$, highlighting in every frame the locations of objects to be edited. 
Such mask can be obtained, for instance, with a pretrained segmentation model~\cite{sam}, visual saliency~\cite{u2net,selfreformer} or by cross-attention with text-prompts.
\subsection{Object-Centric Sampling}
\label{method:oc_sampling}
\input{tables/algo_sampling}
The process of sampling in diffusion models entails gradually transitioning from the initial noise $\z_T \sim \mathcal{N}$ to a sample $\z_0$ from the real data distribution, obtained through a series of denoising steps $p(\z_{t-1} | \z_t)$. Each step involves running a computationally-heavy denoiser (\eg UNet~\cite{unet}), and high-quality generations demand a large number of sampling steps. 
We hereby introduce an efficient object-centric diffusion sampling technique tailored for editing tasks, prioritizing high-quality generation specifically in designated foreground regions.

Our Object-Centric Sampling scheme is described in~\cref{alg:sampling}.
Specifically, we \new{
first consider the foreground mask $m$ and split the latent variable $\z_T$ into foreground and background latents denoted by $\z^f_T$ and $\z^b_T$.
This operation is performed by a generic \emph{gather} function, splitting representation tokens or feature map pixels into two disjoint sets.
Then, instead of performing a single diffusion process, we disentangle the generation of foreground and background regions: such a decoupled scheme allows us to reduce the sampling steps on the latter based on an hyperparameter $\varphi$. 
Once foreground and background latents have been generated, we rely on a \emph{scatter} operation to recompose a full feature map.
Nevertheless, we observed that performing this recomposition step at the end of the whole diffusion process (\ie merging variables $\z^f_0$ and $\z^b_0$) typically results into weak editing performance, in the form of color distribution-shift and boundary artifacts between foreground and background regions.
We therefore introduce two further solutions to this issue.
First, we introduce a normalization step (inspired by batch normalization) applying an affine transformation to the foreground latents, shifting and scaling them to match the mean and standard deviation of background representations.
Even more importantly, we move the scattering of foreground and background latents \emph{within} the sampling process, at a given timestep $T_b$ which is controlled by a hyperparameter $\gamma$.
This choice introduces some diffusion timesteps $t \leq T_b$, that take care of smoothing out such artifacts and seamlessly blend foreground and background latents.
We empirically observed that allocating $\approx \!\! 25\%$ (\ie $\gamma=0.25$) of the sampling steps to this blending stage is usually sufficient to generate spatially and temporally consistent frames.
}
We also note that for localized editing tasks,~\eg shape and attribute manipulations of objects, the background sampling stage can be completely skipped, which results in even faster generation and higher fidelity reconstruction of background regions, as demonstrated in~\cref{fig:ablations_3d.v.2d}.
\subsection{Object-Centric Token Merging}
\label{method:oc_merging}
We further introduce an effective technique that promotes token merging in background areas while discouraging information loss in representations of foreground objects. 
More specifically, whenever applying ToMe, we associate each source token $\x_i$ a binary value $m_i \in \{0,1\}$, obtained by aligning foreground masks to latent resolution, that specifies whether it belongs to the foreground region. Then, we simply account for $m_i$ in computing similarity between the source token $\x_i$ and the destination token $\x_j$, as follows:
\begin{align*}
\eta\text{-Sim}(\x_i, \x_j, m_i) = 
\begin{cases}
&\hspace{-2mm}\text{Sim}(\x_i, \x_j) \quad \text{if} \quad m_i=0;\\
\eta\;\cdot &\hspace{-2mm}\text{Sim}(\x_i, \x_j) \quad \text{if} \quad m_i=1,
\end{cases}
\end{align*}
where $\eta \in [0, 1]$ is a user-defined weighting factor ( $\eta=1$ corresponds to the original ToMe). By reducing the value of $\eta$, we deliberately weaken the similarities of source tokens corresponding to edited objects, therefore reducing their probability of being merged. The behavior of the weighting factor can be appreciated in~\cref{fig:ablations_saliency}: as $\eta$ decreases, the unmerged tokens (in blue) tend to locate more and more on the edited objects, avoiding information loss in their locations.

\paragraph{Merging spatio-temporal token volumes}
Diffusion-based video editing heavily relies on cross-frame attention to increase the temporal-consistency in generated frames~\cite{controlvideo,li2023vidtome}, incurring in severe computational bottlenecks.
The standard ToMe merges tokens only in the spatial dimension, underexploiting the most prominent redundancies in videos, which occur along the time axis.
As a remedy, we apply ToMe \emph{within spatiotemporal volumes}.
This strategy takes advantage of temporal redundancy and allows flexibility in choosing how to trade-off merging spatial vs. temporal information by simply varying the size of the volumes.
\input{figures/ablations_saliency}

%% file: tables/algo_sampling.tex
\begin{algorithm}[t!]
\caption{Object-Centric Sampling}
\label{alg:sampling}
\begin{algorithmic}        
    \Require Number of training steps $T$
    \Comment{Default: 1000}
    \Require Number of inference steps $N$
    \Comment{Default: 20}

    \Require Blending steps ratio $\gamma \in [0, 1]$   
    \Comment{Default: 0.25}
    
    \Require Background acceleration rate $\varphi > 1$       
    \Require Foreground mask $m \in \{0, 1\}^{H
    \times W}$
                
    \State $\Delta T \gets T / N$    
    \Comment{Calculate step size}
    
    \State $T_b \gets \gamma * T$    
    \Comment{Calculate step to start blending}
    
    \State \textit{// Split foreground and background latents}
    \State $\z_T \sim \mathcal{N}(0, I)$    
    \State $\z^f_T \gets \texttt{gather}(\z_T, m)$
    \State $\z^b_T \gets \texttt{gather}(\z_T, 1-m)$
        
    \State \textit{// Sampling the foreground latents at normal rate}
    \For{$t = T \quad ; \quad t > T_b \quad ; \quad t \; = t - \Delta T$}    
        \State $\z^f_{t-1} \sim p(\z^f_{t-1}|\z^f_t)$  
    \EndFor

    \State \textit{// Sampling the background latents at faster rate}
    \For{$t = T \quad ; \quad t > T_b \quad ; \quad t = t - \varphi * \Delta T$}    
        \State $\z^b_{t-1} \sim p(\z^b_{t-1}|\z^b_t)$  
    \EndFor

    \State \textit{// Normalize, merge and continue sampling}
    \new{\State $\bar{\z}_t^f \gets \texttt{normalize}(\z_t^f, \z_t^b)$}
    \State $\z_t \gets \texttt{scatter}(\new{\bar{\z}_t^f}, \z_t^b)$
    \For{$t = T_b \quad ; \quad t > 0 \quad ; \quad t \; = t - \Delta T$}    
        \State $\z_{t-1} \sim p(\z_{t-1}|\z_t)$  
    \EndFor
\end{algorithmic}
\end{algorithm}

%% file: figures/ablations_saliency.tex
\begin{figure}[t]
\centering
\hspace{-3mm}
\includegraphics[width=0.6\linewidth]{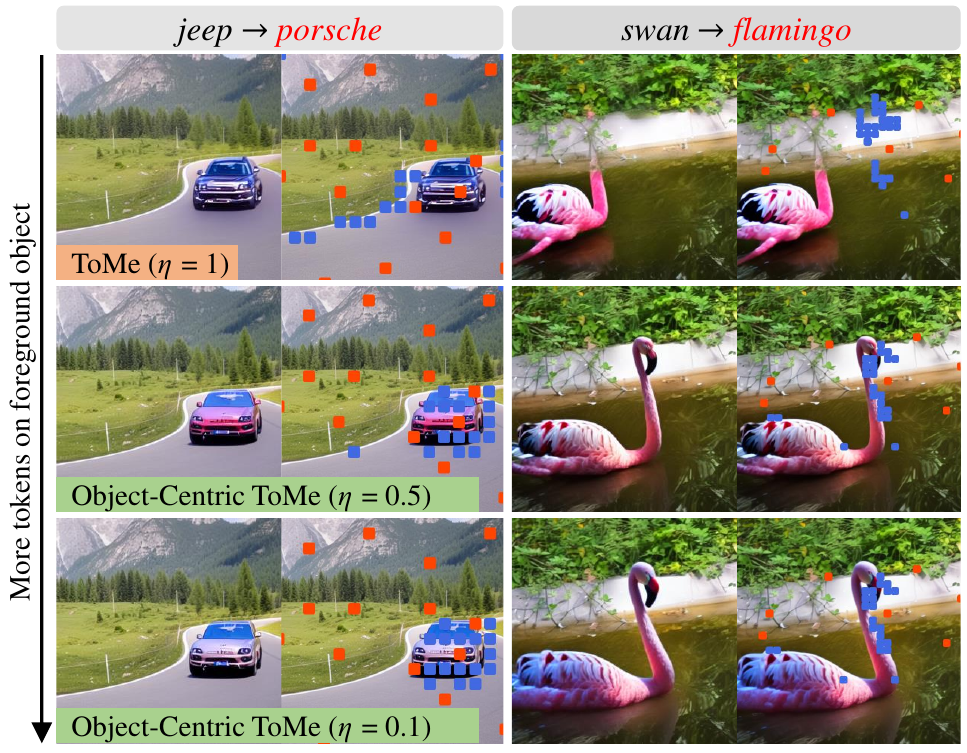}
\caption{
\textbf{Object-Centric Token Merging:} 
By artificially down-weighting the similarities of source tokens of foreground objects, we accumulate in their locations tokens that are left unmerged (in \textcolor{blue}{blue}). 
Destination tokens (in \textcolor{red}{red}) are still sampled randomly within a grid, preserving some background information.
Merged source tokens (not represented for avoiding cluttering) will come from the background.
}
\label{fig:ablations_saliency}
\end{figure}

%% file: texts/5_experiments.tex
\input{figures/fatezero_comparisons}
\section{Experiments}
We test OCD in the context of two families of video editing models.
Specifically, we look into inversion-based video editing, where we rely on FateZero~\cite{fatezero} as base model, and control-signal-based architectures, for which we optimize a ControlVideo~\cite{controlvideo} baseline.
To provide fair comparisons in each setting, we rely on checkpoints and configurations provided by the corresponding authors.
We evaluate each model using a benchmark composed of DAVIS~\cite{davis2017} video sequences and edit prompts utilized in the original baseline methods.
\\\\
\new{
\paragraph{Implementation details}
For both video editing models, we pad the mask $m$ to a rectangular shape and implement the gather operation in~\cref{alg:sampling} as a crop around the foreground object.
Although potentially suboptimal for non-rectangular objects, this strategy proves effective in practice (\cref{tab:ablation_blending_real}) and enables the use of regular dense convolutions within UNet, which are typically highly optimized and even faster than sparse counterparts.
Such an approximation would however not be necessary for other convolution-free editing pipelines, \eg fully based on transformer architectures~\cite{peebles2023dit}.
As common in the literature and because we target zero-shot video editing, we optimize some of the hyperparameters (\eg blending steps ratio $\gamma$, similarity re-weighting factor $\eta$) per-sequence for FateZero.
For ControlVideo, we adopt the original hyperparameters reported in the paper.
$\eta$ is selected in $\{0.1, 0.5, 0.9\}$ based on heuristics: a rule-of-thumb, is to use a lower value (preserving more foreground), when the object has complex motion or texture (i.e., it is challenging to reconstruct).
As for the saliency masks, we utilize the segmentation masks available within the DAVIS dataset.
}
We refer the reader to \supp{supplementary materials} for more-detailed setup.

\paragraph{Evaluation metrics}
For evaluating editing quality, we rely on fidelity metrics such as Temporal-Consistency of CLIP embeddings (Tem-con) and average CLIP-score aggregated over all sequence-prompt pairs.
To report latency, we measure the average wall-clock time to edit a video on a single V100 GPU.
\subsection{Inversion-based Video Editing}
Following the benchmark in~\cite{fatezero}, we report a quantitative comparison based on 9 sequence-prompt pairs.
We include inversion-based video models such as Tune-A-Video + DDIM~\cite{tuneavideo,ddim}, TokenFlow + PnP~\cite{tokenflow,plugandplay} and \new{VidToMe~\cite{li2023vidtome}} in the evaluation.
We also show results for a frame-based editing model, namely Framewise Null + p2p~\cite{nulltext,p2p}.
We finally report results of the Optimized-FateZero baseline, obtained by the off-the-shelf accelerations in \cref{sec:offtheshelf}.
\input{tables/metrics_fatezero}

\paragraph{Main results} 
A qualitative comparison of OCD and state-of-the-art models is reported in~\cref{fig:fatezero_comparison}, where our model enjoys the lowest latency, while ensuring a comparable editing quality.
We report fidelity and latency measurements in~\cref{tab:sota_fatezero}, where OCD achieves remarkable latency gains of 10$\times$ w.r.t. FateZero, and proves significantly faster than other state-of-the-art methods with comparable fidelity metrics.
Although Optimized-FateZero is also very fast, we observe via visual assessment that its editing quality can be suboptimal, even though highly localized geneation artifacts might be overlooked by fidelity metrics (for clear examples of this, we refer to the ablation in~\cref{fig:ablations_3d.v.2d}).
\subsection{ControlNet-based Video Editing}
We follow the evaluation benchmark in~\cite{controlvideo} for evaluating ControlNet-based algorithms.
The comparison comprises 125 sequence-prompt pairs from DAVIS (detailed in the \supp{supplement}), for which per-frame depth maps are extracted using~\cite{ranftl2020towards} and used as control signal.
Besides our main baseline ControlVideo~\cite{controlvideo}, we include Text2Video-Zero~\cite{text2video-zero} in the comparisons.
Once again, we will also report the performances of off-the-shelf optimizations (Optimized-ControlVideo).

\paragraph{Main results} 
\cref{fig:qualitative_controlvideo} illustrates some visual examples.
As the figure shows, our proposal shows a significant $6\times$ speed-up over the baseline, obtaining a comparable generation quality.
We also report a quantitative comparison of ControlNet-based methods in~\cref{tab:sota_controlvideo}, observing similar outcomes.
We notice that Text2Video-Zero is slightly faster than our method, due to its sparse instead of dense cross-frame attention.
However, for the same reason it underperforms in terms of temporal-consistency among generated frames. 
We refer the reader to the \supp{supplementary material} for additional qualitative comparison.
\input{figures/controlvideo_table_and_qualitative}
\subsection{Analysis}
\paragraph{Ablation study} 
\input{figures/ablation_study_table_and_qualitative}
We ablate our main contributions within the FateZero base model. 
Specifically, we illustrate the impact of our proposals in~\tref{tab:ablation_fatezero}, where we report Tem-con and latency as metrics, and in~\fref{fig:ablations_3d.v.2d}.
Although off-the-shelf optimizations may grant reasonable speed-ups, we found that for extreme token reduction rates their editing results can suffer from generation artifacts.
As the figure shows, the addition of Object-Centric ToMe can easily fix most defects on foreground regions by forcing ToMe to operate on other areas, a benefit that comes without major impacts on latency.
Finally, Object-Centric Sampling enables significant latency savings performing most diffusion iterations on foreground areas only.
Somewhat surprisingly, we observe that Object-Centric Sampling helps the generation of background areas too: this is due to the fact that, as we run fewer denoising steps, its representations remain closer to the ones of the original video (resulting from inversion).

\input{figures/saliency_source}
\paragraph{Impact of saliency mask} \new{We compare OCD by using saliency masks from three sources: i) human-labeled masks, ii) predictions from a state-of-the-art segmentation models (Grounded-SAM~\cite{ren2024groundedsam}), and iii) cross-attention maps w.r.t.~text prompts (available within the model itself). We did not observe meaningful differences either in generations (see \cref{fig:saliency_quality} for an example) or in aggregated fidelity metrics (\cref{tab:saliency_source}), suggesting that OCD is robust to the source of saliency.}

\paragraph{Object-Centric Sampling at different object sizes}
The latency impact of Object-Centric Sampling depends on the size of the foreground areas: in the presence of very small edits it can allow significant savings, whereas for prominent objects its editing cost would be only mildly-affected.
We study this behavior in~\cref{tab:ablation_blending_real}, where we divide 18 sequence-prompt pairs in three sets of 6 pairs each, 
\new{depending on the number of foreground pixels (or foreground ratio $\Delta$): Large  ($\Delta>20\%$), Medium ($\Delta\in[10,20]\%$) and Small ($\Delta<10\%$). 
}
As we show in~\cref{tab:ablation_blending_real}, in the absence of Object-Centric Sampling, the latency of video editing does not depend on the size of the foreground objects.
\new{However, although we remark that our proposal reduces latency for all object sizes, we observe smaller gains on large objects, as the foreground area covers most of the frame resolution, matching Object-Centric Sampling with a regular diffusion process operating on the whole frame.
Savings gradually improve as we go towards medium and small sized objects, showing additional speed-ups of 1.3$\times$ and 1.8$\times$ respectively.}
\input{tables/ablation_fatezero_blending_real}

\paragraph{Limitations}
We highlight some potential limitations of our model to be tackled by future work.
Although Object-Centric Diffusion represent valuable ideas in general, they are particularly effective for local changes to some peculiar objects, and are slightly less-suitable for global editing (\eg changing the style/textures of a video altogether).
Moreover, as most zero-shot methods, in order to achieve the best trade-off between fidelity and latency, our framework still requires to search the best hyperparameters per-sequence.

%% file: figures/fatezero_comparisons.tex
\begin{figure*}[t]
\centering
\includegraphics[width=1\linewidth]{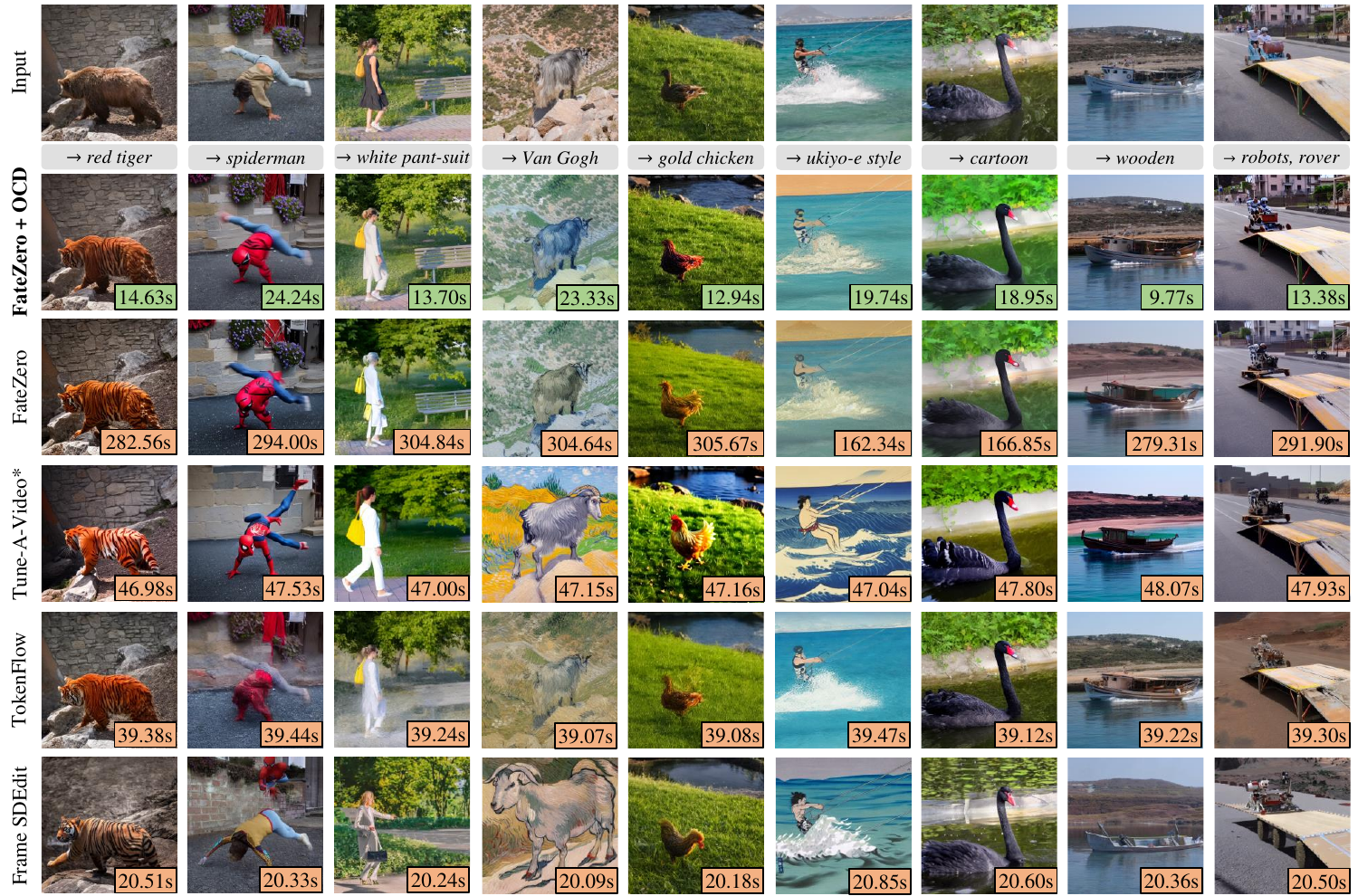}
\caption{
\textbf{Qualitative comparison with the sota editing methods:} 
OCD yields significantly-faster generations over the baseline we build on-top of (i.e., FateZero \cite{fatezero}) and other state-of-the-art methods, without sacrificing quality.
Tune-A-Video~\cite{tuneavideo} is finetuned on each sequence (denoted with *, finetuning time not included in latency).
}
\label{fig:fatezero_comparison}
\end{figure*}

%% file: tables/metrics_fatezero.tex
\begin{table}[t!] %
\centering
\caption{
\textbf{Quantitative results in inversion-based pipelines:} 
Our method achieves significant speed-up compared to the baseline and other the state-of-the-art methods (either video or framewise), without sacrificing generation quality.
}
\tablestyle{1.8pt}{1.}
\resizebox{0.8\columnwidth}{!}{
\def\arraystretch{1.2}%
\begin{tabular}{lccrr}
\toprule
\multirow{2}{*}{Model} & \multirow{2}{*}{Tem-con $\uparrow$} & \multirow{2}{*}{Cl-score $\uparrow$} & \multicolumn{2}{c}{Latency (s) $\downarrow$} \\
\cmidrule(l{0.5mm}){4-5}
& & & Inversion & Generation \\ %
\midrule
Framewise Null + p2p~\cite{nulltext,p2p} & 0.896 & 0.318 & 1210.60 & 130.24 \\
Tune-A-Video + DDIM~\cite{tuneavideo,ddim} & 0.970 & 0.335 & 16.50 & 33.09 \\ %
TokenFlow + PnP~\cite{tokenflow, plugandplay} & 0.970 & 0.327 & 10.56 & 28.54 \\
VidToMe~\cite{li2023vidtome} & 0.961 & 0.326 & 9.28 & 39.83 \\
FateZero~\cite{fatezero} & 0.961 & 0.344 & 135.80 & 41.34 \\ %
\rowcolor{row} Optimized-FateZero & 0.966 & 0.334 & 9.54 & 10.14 \\ %
\rowcolor{row} \quad + Object-Centric Diffusion & 0.967 & 0.331 & \textbf{8.22} & \textbf{9.29} \\ %
\bottomrule
\end{tabular}}
\label{tab:sota_fatezero}
\end{table}

%% file: figures/controlvideo_table_and_qualitative.tex
\bgroup
\newcommand{\canny}[1]{}
\begin{figure}[t]
\centering\resizebox{0.7\columnwidth}{!}{
\begin{tabular}{lccr}
\toprule
Model & Tem-con $\uparrow$ & Cl-score $\uparrow$ & Latency (s) $\downarrow$ \\
\midrule
Text2Video-Zero~\cite{text2video-zero} & 0.960 \canny{(0.952)} & 0.317 \canny{(0.307)} & \textbf{23.46} \\
ControlVideo~\cite{controlvideo} & 0.972 \canny{(0.968)} & 0.318 \canny{(0.308)} & 152.64 \\
\rowcolor{row} Optimized-ControlVideo & 0.978 \canny{(0.972)} & 0.314 \canny{(0.303)} & 31.12\\
\rowcolor{row} \quad + Object-Centric Diffusion & 0.977 \canny{(0.967)} & 0.313 \canny{(0.302)} & 25.21 \\
\bottomrule
\end{tabular}}%

\includegraphics[width=0.55\columnwidth]{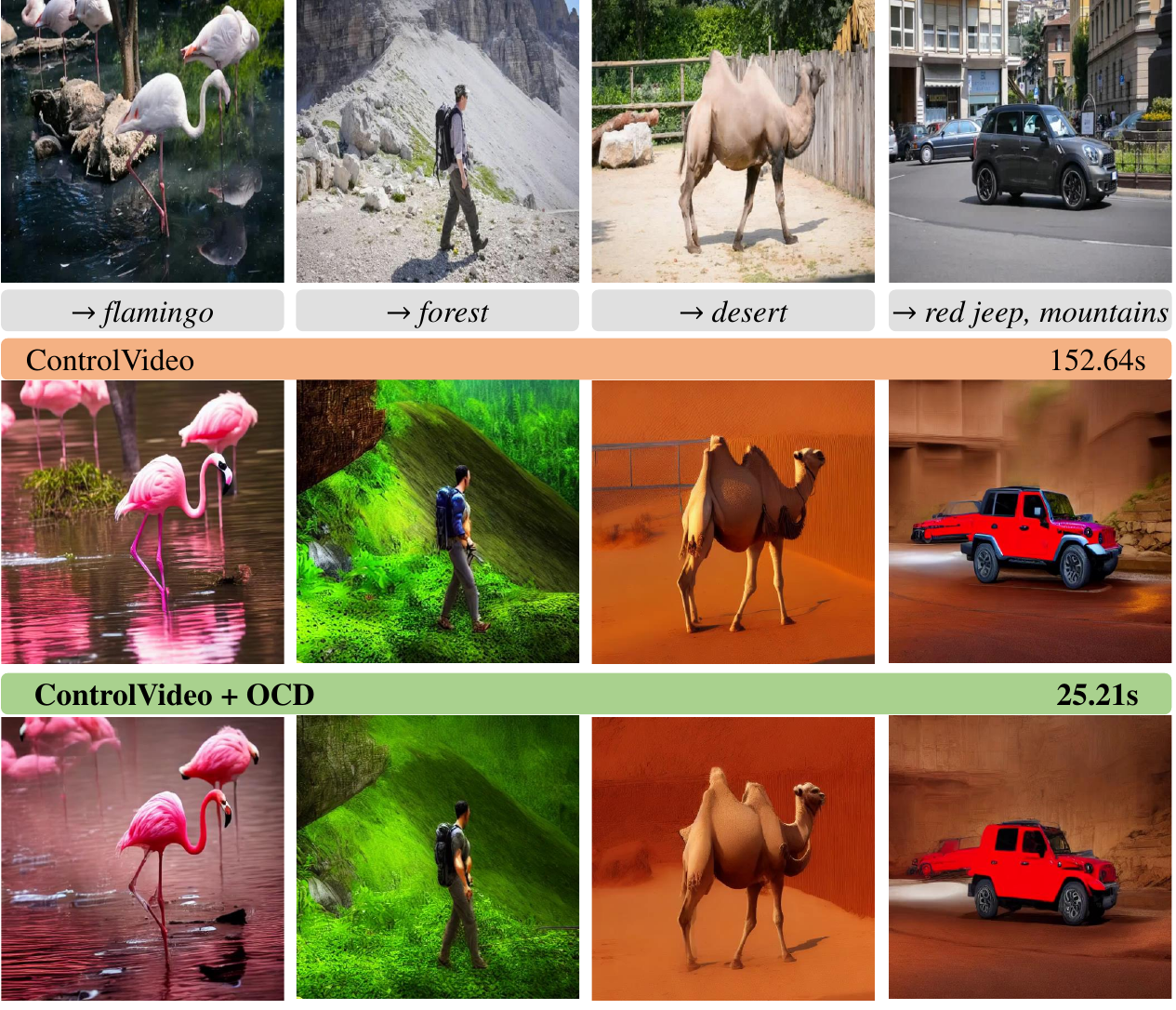}
\captionlistentry[table]{}
\label{tab:sota_controlvideo}
\captionsetup{labelformat=andfigure}
\caption{
\textbf{Comparison with ControlNet-based pipelines:} we report both quantitative and qualitative results based on  depth conditioning. With comparable generation quality, our method achieves a 6$\times$ speed-up compared to the baseline.
}
\label{fig:qualitative_controlvideo}
\end{figure}
\egroup

%% file: figures/ablation_study_table_and_qualitative.tex
\begin{figure}[tbh]
\centering
\hspace{-4mm}
\resizebox{0.65\columnwidth}{!}{
\begin{tabular}{lcrr}
\toprule
\multirow{2}{*}{Component} & \multirow{2}{*}{Tem-con $\uparrow$} & \multicolumn{2}{c}{Latency (s) $\downarrow$} \\
\cmidrule(l{0.5mm}){3-4}
& & Inversion & Generation \\
\midrule
Optimized-FateZero & 0.966 & 8.90 & 10.38 \\
\quad + Object-Centric ToMe & 0.967 & 9.05 & 10.54 \\
\qquad + Object-Centric Sampling & 0.966 & \textbf{7.59} & \textbf{9.58} \\
\bottomrule
\end{tabular}}
\includegraphics[width=0.6\linewidth]{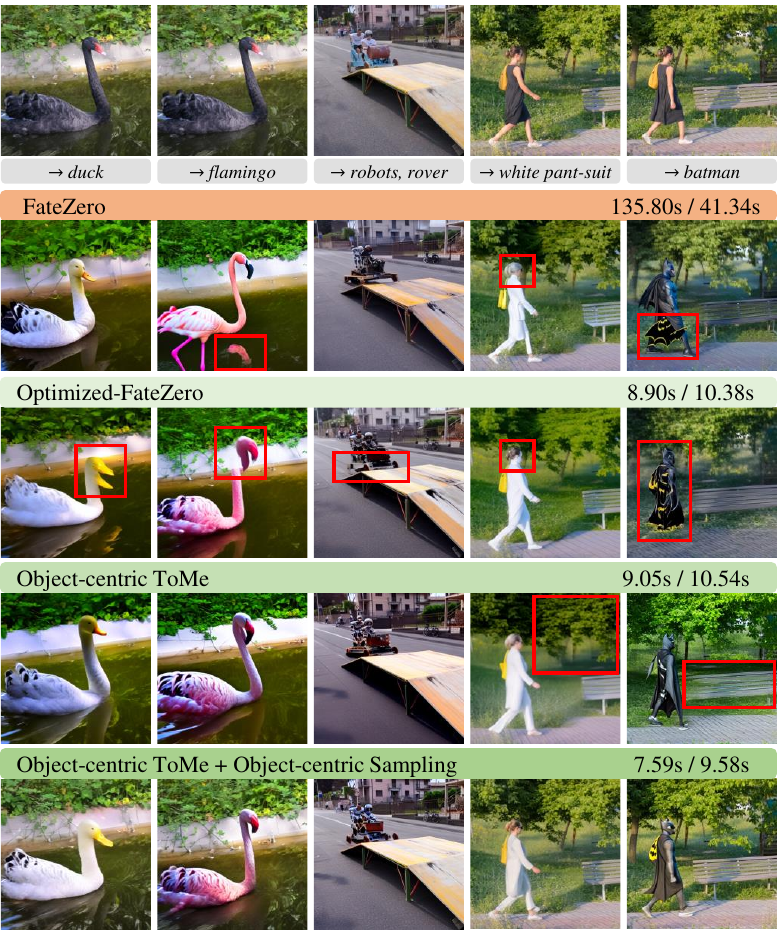}
\captionlistentry[table]{}
\label{tab:ablation_fatezero}
\captionsetup{labelformat=andfigure}
\caption{
\textbf{Ablation of OCD components:} 
Object-Centric ToMe improves temporal-consistency and fidelity without sacrificing latency. Object-Centric Sampling further improves latency. We highlight artifacts with a red outline.} 
\label{fig:ablations_3d.v.2d}
\end{figure}

%% file: figures/saliency_source.tex
\bgroup
\newcommand{\canny}[1]{}
\begin{figure}[t]
\centering\resizebox{0.45\columnwidth}{!}{
\begin{tabular}{lcc}
\toprule
Saliency mask & Tem-con $\uparrow$ & Cl-score $\uparrow$ \\
\midrule
GT mask & 0.967 & 0.331 \\
Grounded SAM \cite{ren2024groundedsam} & 0.967 & 0.329 \\
Cross-attention maps & 0.966 & 0.329 \\
\bottomrule
\end{tabular}}
\includegraphics[width=0.85\linewidth]{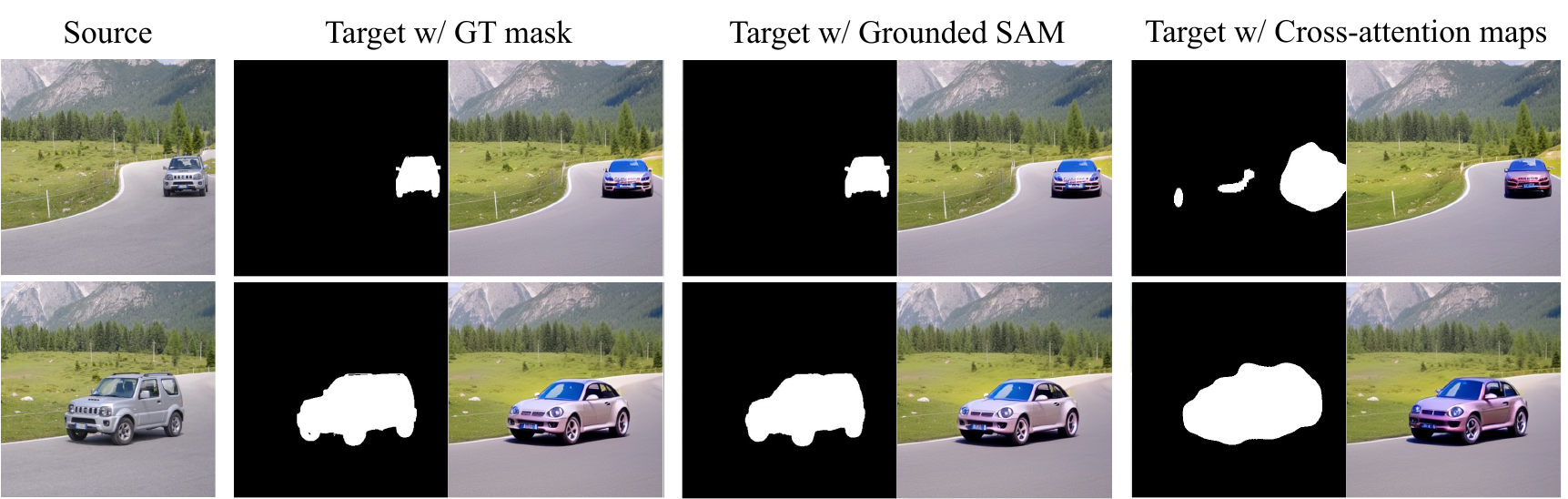}
\captionlistentry[table]{}
\label{tab:saliency_source}
\captionsetup{labelformat=andfigure}
\caption{
\new{\textbf{Edit quality w.r.t. saliency masks:} 
OCD generates faithful edits even with coarse masks, showing its resiliency to suboptimal saliency sources.}}
\label{fig:saliency_quality}
\end{figure}
\egroup

%% file: tables/ablation_fatezero_blending_real.tex
\begin{table}[t] %
\centering
\caption{
\new{\textbf{Impact of Object-Centric Sampling at different object sizes.} 
We achieve more latency savings with smaller foreground objects without sacrificing the generation quality.
Latency reductions are shown in \textcolor{gray}{gray}.}
}
\tablestyle{1.8pt}{1.}
\resizebox{0.9\columnwidth}{!}{
\begin{tabular}{llccccrr}
\toprule
\multirow{2.5}{*}{Object} & \multirow{2.5}{*}{Avg.~\#pixel} & \multicolumn{3}{c}{w/o Obj-Cen. Sampling$\;\;$} & \multicolumn{3}{c}{\cellcolor{row}w/ Obj-Cen. Sampling}\\
\cmidrule(l{0.5mm}r{1.5mm}){3-5}
\cmidrule(l{0mm}){6-8}
& & \multirow{2}{*}{Tem-con $\uparrow$} & \multicolumn{2}{c}{Latency (s) $\downarrow$} & \multirow{2}{*}{Tem-con $\uparrow$$\;$} & \multicolumn{2}{c}{Latency (s) $\downarrow$} \\
\cmidrule(l{1mm}r{1mm}){4-5}
\cmidrule(l{1mm}){7-8}
\multirow{-3.2}{*}{size} & \multirow{-3.2}{*}{(fg. ratio $\Delta$)$\;$} & & \multicolumn{1}{c}{Inversion} & \multicolumn{1}{c}{Generation} &  & \multicolumn{1}{c}{Inversion} & \multicolumn{1}{c}{Generation} \\
\midrule
Large & 53.5k (20.4\%) & 0.976 & $\;$12.70$\;$ & $\;\;$11.08$\;\;$ & 0.967 & 9.77 \textcolor{gray}{(2.93$\downarrow$)} & 9.65 \textcolor{gray}{(1.43$\downarrow$)} \\
Medium$\;$ & 35.2k (13.4\%) & 0.956 & 13.09 & 11.08 & 0.953 & 9.97 \textcolor{gray}{(3.12$\downarrow$)} & 8.82 \textcolor{gray}{(2.26$\downarrow$)} \\
Small & 17.7k (6.7\%) & 0.953 & 13.02 & 10.77 & 0.948 & \textbf{7.25} \textcolor{gray}{(5.77$\downarrow$)} & \textbf{6.22} \textcolor{gray}{(4.55$\downarrow$)} \\
\bottomrule
\end{tabular}}
\label{tab:ablation_blending_real}
\end{table}

%% file: texts/6_conclusion.tex
\section{Conclusion}
In this paper, we introduced solutions for speeding up diffusion-based video editing.
In this respect, we first presented an analysis of sources of latency in inversion-based models, and we identified and adopted some off-the-shelf techniques such as fast sampling and Token Merging that, when properly modified for the task, bring significant cost reduction with only slight quality degradation.
Furthermore, motivated by the fact that video editing typically requires modifications to specific objects, we introduce Object-Centric Diffusion, comprising techniques for \emph{i)} encouraging the merging of tokens in background regions and \emph{ii)} limiting most of the diffusion sampling steps on foreground areas.
Our solutions, which fix generation artifacts on foreground objects and further reduce editing latency, are validated on inversion-based and ControlNet-based models, by achieving 10$\times$ and 6$\times$ faster edits for comparable quality.

%% file: texts/supplementary/A_discussion.tex
\renewcommand{\thetable}{A.\arabic{table}}
\renewcommand{\thefigure}{A.\arabic{figure}}
\renewcommand{\thesection}{A}
\renewcommand{\thesubsection}{A.\arabic{subsection}}
\setcounter{table}{0}
\setcounter{figure}{0}

\new{This supplementary is organized into two sections. First, in \sref{sec:sup_A}, we present additional discussion on off-the-shelf-optimizations and benchmark settings. In \sref{sec:sup_B}, we present additional results, both qualitative and quantitative. Project page: \href{https://qualcomm-ai-research.github.io/object-centric-diffusion}{\texttt{qualcomm-ai-research.github.io/object-centric-diffusion}}.
}

\section{Additional Discussion}
\label{sec:sup_A}
\subsection{Off-the-shelf optimizations of ToMe}

\paragraph{Pairing token locations from inversion}
Many inversion-based image/video editing pipelines rely on sharing attention maps between inversion and generation stages (\eg FateZero~\cite{fatezero}, Plug-and-Play~\cite{plugandplay}).
As such, when applying ToMe \cite{bolya2022token,bolya2023token}, it is important that locations of destination (\textit{dst}) and unmerged (\textit{unm}) tokens are paired in the two stages, at each corresponding attention layer and diffusion step.
If that is not the case, tokens or attention maps coming from inversion are not compatible with the ones available at generation time.
In practice, we compute which tokens to be merged during inversion, and merge the tokens at the same locations in generation attention maps. \new{By doing so, we make sure the tokens that remain after merging correspond to the same locations (or, token indices), and hence, the attention maps from inversion and generation can rightly be fused.}
We found this strategy to be of primary importance, as testified by Fig.~3 (d-e) in the main paper.

\paragraph{Re-sampling destination tokens per-frame}
ToMe for stable diffusion~\cite{bolya2023token} samples \textit{dst} tokens randomly \new{in a single image}. 
When extending this to multiple frames, we initially sample the same random locations in each frame, finding this strategy to be sub-optimal. Instead, if we re-sample different random locations in each frame (or, in each temporal window in our \new{spatio-temporal} implementation), it allows us to preserve different information \new{in each frame (or, window)} after merging. We found this to be beneficial, especially at higher merging rates (\eg see Fig.~3 (e to f) in the main paper).

\paragraph{How to search for destination match}
In the original ToMe for stable diffusion~\cite{bolya2023token}, for each source (\textit{src}) token, we search its corresponding match within a pool of \textit{dst} tokens coming from the full spatial extent of the given image ($H\times W$).
The naive transposition of this strategy to our video use-case allows, for any \textit{src} token, its candidate match to be searched within a pool of \textit{dst} tokens coming from the full spatio-temporal extent of the video ($T\times H\times W$).
We find that this strategy for searching  \textit{dst} match, named hereby \textit{Temporally-Global Search}, can lead to generation artifacts.
Differently, we consider restricting the temporal extent of the \textit{dst} pool to be the same temporal-window ($s_t\times H\times W$) as the \textit{src} token, as in our \textit{Temporally-Windowed Search}.
As shown in~\fref{fig:ablations_3dw.v.3dg}, the latter gives better reconstructions in general, whilst allowing more flexibility to control where merges are happening\new{, temporally}.
This way, the user can also better trade-off the temporal-redundancy, smoothness and consistency by controlling the spatio-temporal window size.
\input{figures/ablations_3dw.v.3dg}

\paragraph{Merging queries, keys or values?}
In our early experiments, we consider applying ToMe to all queries (with unmerging, as in~\cite{bolya2023token}), keys and values.
We however find that, with extreme reduction rates, merging queries can easily break the reconstructions.
As such, we limit ToMe to operate on keys and values only.
We also observe that in dense cross-frame attention modules, merging queries only provide a slight latency reduction.

\paragraph{Capped merging in low-res UNet stages}
As observed in~\cite{bolya2023token}, the high resolution UNet \cite{unet} stages are the most expensive in terms  of self-attention (or, cross-frame attention) modules, and the ones that can benefit the most by applying ToMe.
Contrarily to the original formulation which does not optimize low-resolution layers, we do apply ToMe in all layers as we observe it has a meaningful impact on latency.
We however cap the minimum \#tokens \new{preserved after merging} in low-resolution layers, in order to avoid degenerate bottlenecks (\eg collapsing to a single representation). Specifically, we maintain at-least 4 and 16 tokens per-frame after merging at $8\times8$ and $16\times16$ resolutions, respectively.
\subsection{Benchmark settings}
\paragraph{Evaluation metrics} We consider two metrics for quantitative evaluation: CLIP-score and Temporal-consistency, similar to prior work~\cite{fatezero,controlvideo}.
CLIP-score is computed as the cosine similarity between CLIP \cite{clip} visual embedding of each frame and CLIP text embedding of the corresponding edit prompt, aggregated over all frames and sequences. It measures the semantic fidelity of the generated video.
Temporal-consistency is computed as the cosine similarity between the CLIP visual embeddings of each consecutive pairs of frames, aggregated over all pairs and sequences. It conveys the visual quality of the generated video, measuring how temporally coherent frames are.
We highlight that, despite their use is due in fair comparisons due to their popularity, both these fidelity metrics are far from perfect.
For instance, we find the CLIP score to be sensitive to global semantic discrepancies, yet it often overlooks generation artifacts and smaller pixel-level details.
Furthermore, Temporal-Consistency can be simply exploited by a fake video repeating a frame over time.
For these reasons, extensive visual comparisons are still required to assess different models, and future research should be encouraged towards more informative quantitative protocols for video editing.
\paragraph{Sequence-prompt pairs} 
We present the sequence-prompt pairs considered in our evaluation of inversion-based pipelines in \tref{tab:prompts}. Most sequences here are from DAVIS \cite{davis2017} dataset, with the exception of a few in-the-wild videos introduced in \cite{fatezero}. The Benchmark setting corresponds to the original quantitative evaluation of FateZero \cite{fatezero}, which includes 9 sequence-prompt pairs. We also present the sequence-prompt pairs used to evaluate our Object-Centric Sampling (see \new{Table 5} in the main paper), categorized based on the foreground object size: Large, Medium and Small. 
In \tref{tab:prompts_control}, we show the 125 sequence-prompt pairs used in ControlNet-based pipelines, provided by the authors of ControlVideo \cite{controlvideo}.
\clearpage
\input{tables/supp_prompts_fatezero}
\clearpage
\input{tables/supp_prompts_controlvideo}
\clearpage

%% file: figures/ablations_3dw.v.3dg.tex
\begin{figure}[t!]
    \centering
    \includegraphics[width=0.8\linewidth]{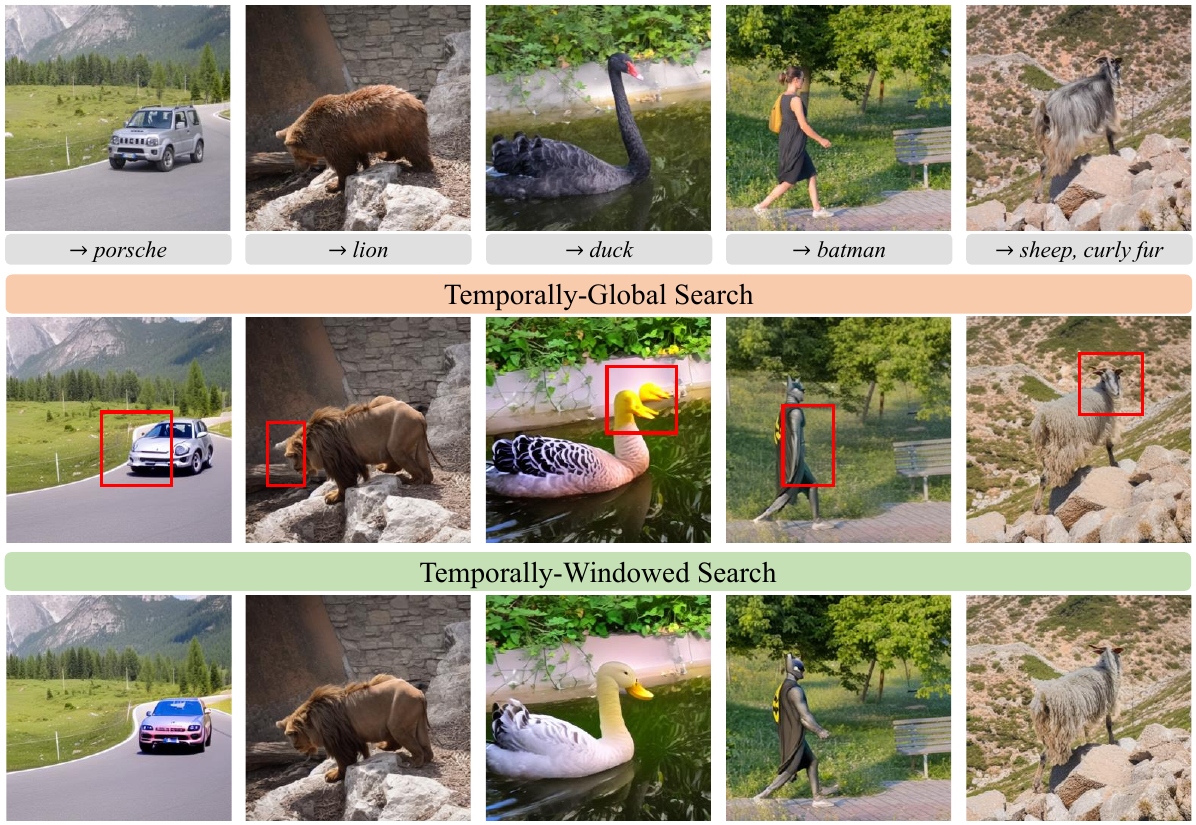}
    \caption{\textbf{Qualitative ablation on searching \textit{dst} match} Each \textit{src} token may search for its corresponding match within a pool of \textit{dst} tokens. This pool can be comprised of either the whole spatio-temporal extent (i.e., all frames) as in \textit{Temporally-Global Search}, or only the same temporal window as the corresponding \textit{src} token as in \textit{Temporally-Windowed Search}. Among these two strategies, the latter allows more flexibility, providing more-consistent generations with better fidelity.}
    \label{fig:ablations_3dw.v.3dg}
\end{figure}

%% file: tables/supp_prompts_fatezero.tex
\setcounter{table}{0}
\begin{table*}[h] %
\centering
\caption{\textbf{Sequence-prompt pairs used to evaluate inversion-based pipelines}: Most sequences here are from DAVIS \cite{davis2017}, except for a few in-the-wild videos used in \cite{fatezero}. The Benchmark pairs correspond to the original FateZero \cite{fatezero} quantitative evaluation setting. \new{We also show the sequence-prompt pairs used to evaluate our Object-Centric Sampling, separately for Large, Medium and Small objects.} 
}
\tablestyle{1.8pt}{1.}
\resizebox{0.85\textwidth}{!}{
\scriptsize
\begin{tabular}{m{0.02\textwidth} m{0.14\textwidth} m{0.3\textwidth} m{0.53\textwidth}}
\cmidrule[0.75pt]{1-4}
& \multirow{1}{*}{Sequence} & Source prompt & $\;\;\;$Target prompts \\
\cmidrule{1-4}
\cellcolor{white} \multirow{23}{*}{\rotatebox[origin=c]{90}{FateZero \cite{fatezero} Benchmark}} & \cellcolor{rowx} blackswan & \cellcolor{rowx} a black swan with a red beak swimming in a river near a wall and bushes. & \cellcolor{rowx}
\begin{itemize}[topsep=0pt]
\item a \pr{white duck} with a \pr{yellow} beak swimming in a river near a wall and bushes.
\item a \pr{pink flamingo} with a red beak \pr{walking} in a river near a wall and bushes.
\item  a \pr{Swarovski crystal} swan with a red beak swimming in a river near a wall and bushes.
\item  \pr{cartoon photo of} a black swan with a red beak swimming in a river near a wall and bushes. 
\end{itemize}\\
& car-turn & a silver jeep driving down a curvy road in the countryside. & 
\begin{itemize}[topsep=0pt]
\item a \pr{Porsche car} driving down a curvy road in the countryside.
\item \pr{watercolor painting of} a silver jeep driving down a curvy road in the countryside.
\end{itemize}\\
& \cellcolor{rowx} kite-surf & \cellcolor{rowx} a man with round helmet surfing on a white wave in blue ocean with a rope. & \cellcolor{rowx}
\begin{itemize}[topsep=0pt]
\item a man with round helmet surfing on a white wave in blue ocean with a rope \pr{in the Ukiyo-e style painting}.
\end{itemize}\\
& train $\;\;\;\;\;\;\;\;$(in-the-wild) & a train traveling down tracks next to a forest filled with trees and flowers and a man on the side of the track. &
\begin{itemize}[topsep=0pt]
\item a train traveling down tracks next to a forest filled with trees and flowers and a man on the side of the track \pr{in Makoto Shinkai style}.
\end{itemize}\\
& \cellcolor{rowx} rabbit $\;\;\;\;\;\;\;$(in-the-wild) & \cellcolor{rowx} a rabbit is eating a watermelon. & \cellcolor{rowx}
\begin{itemize}[topsep=0pt]
\item \pr{pokemon cartoon of} a rabbit is eating a watermelon.
\end{itemize}\\
\cmidrule{1-4}
\cellcolor{white} \multirow{10}{*}{\rotatebox[origin=c]{90}{Large object}} & \cellcolor{rowx} blackswan & \cellcolor{rowx} a black swan with a red beak swimming in a river near a wall and bushes. & \cellcolor{rowx}
\begin{itemize}[topsep=0pt]
\item a \pr{white duck} with a \pr{yellow} beak swimming in a river near a wall and bushes.
\item a \pr{pink flamingo} with a red beak \pr{walking} in a river near a wall and bushes.
\item a \pr{Swarovski crystal swan} with a red beak swimming in a river near a wall and bushes.
\end{itemize}\\
& bear & a brown bear walking on the rock against a wall. &
\begin{itemize}[topsep=0pt]
\item a \pr{red tiger} walking on the rock against a wall.
\item a \pr{yellow leopard} walking on the rock against a wall.
\end{itemize}\\
\cmidrule{1-4}
\cellcolor{white} \multirow{14}{*}{\rotatebox[origin=c]{90}{Medium object}} & \cellcolor{rowx} breakdance & \cellcolor{rowx} a man wearing brown tshirt and jeans doing a breakdance flare on gravel. & \cellcolor{rowx} 
\begin{itemize}[topsep=0pt]
\item a \pr{woman with long-hair} wearing \pr{green-sweater} and jeans doing a breakdance flare on gravel.
\item a \pr{spiderman} wearing \pr{red-blue spidersuit} doing a breakdance flare on gravel.
\item a \pr{chimpanzee} wearing a \pr{black} jeans doing a breakdance flare on gravel.
\end{itemize}\\
& boat & a white color metal boat cruising in a lake near coast. &
\begin{itemize}[topsep=0pt]
\item a \pr{heavily-rusted} metal boat cruising in a lake near coast.
\item a \pr{light-brown} color \pr{wooden} boat cruising in a lake near coast.
\end{itemize}\\
& \cellcolor{rowx} car-turn & \cellcolor{rowx} a silver jeep driving down a curvy road in the countryside. & \cellcolor{rowx}
\begin{itemize}[topsep=0pt]
\item a \pr{Porsche car} driving down a curvy road in the countryside.
\end{itemize}\\

\cmidrule{1-4}
\cellcolor{white} \multirow{15}{*}{\rotatebox[origin=c]{90}{Small object}} & mallard & a brown mallard running on grass land close to a lake. &
\begin{itemize}[topsep=0pt]
\item a \pr{white duck} running on grass land close to a lake.
\item a \pr{golden chicken} running on grass land close to a lake.
\item a \pr{gray goose} running on grass land close to a lake.
\end{itemize}\\
& \cellcolor{rowx} lucia & \cellcolor{rowx} a woman wearing a black dress with yellow handbag walking on a pavement. & \cellcolor{rowx} 
\begin{itemize}[topsep=0pt]
\item a woman wearing a \pr{white pant-suit} with yellow handbag walking on a pavement.
\item a woman wearing a black dress \pr{and a hat} with \pr{red} handbag walking on a pavement.
\item a \pr{batman} wearing a black \pr{bat-suit} walking on a pavement.
\end{itemize}\\

& soapbox & two men driving a soapbox over a ramp. & 
\begin{itemize}[topsep=0pt]
\item two \pr{robots} driving a \pr{mars-rover} over a ramp.
\end{itemize}\\
\cmidrule[0.75pt]{1-4}
\end{tabular}}
\label{tab:prompts}
\end{table*}

%% file: tables/supp_prompts_controlvideo.tex
\begin{table*}[h!] %
\centering
\caption{\textbf{Sequence-prompt pairs used to evaluate ControlNet-based pipelines}: All sequences are from DAVIS \cite{davis2017}. These pairs correspond to the original ControlVideo \cite{controlvideo} quantitative evaluation setting. [continued...]}
\tablestyle{1.8pt}{1.}
\resizebox{0.9\textwidth}{!}{
\scriptsize
\begin{tabular}{m{0.14\textwidth} m{0.20\textwidth} m{0.78\textwidth}}
\cmidrule[0.75pt]{1-3}
\multirow{1}{*}{Sequence} & Source prompt & $\;\;\;$Target prompts \\
\cmidrule{1-3}

\rowcolor{rowx} blackswan &  a black swan moving on the lake. &
\begin{itemize}[topsep=0pt]
\item A black swan moving on the lake \item A white swan moving on the lake. \item A white swan moving on the lake, cartoon style. \item A crochet black swan swims in a pond with rocks and vegetation. \item A yellow duck moving on the river, anime style.
\end{itemize}\\

boat & a boat moves in the river. &
\begin{itemize}[topsep=0pt]
\item A sleek boat glides effortlessly through the shimmering river, van gogh style. \item A majestic boat sails gracefully down the winding river. \item A colorful boat drifts leisurely along the peaceful river. \item A speedy boat races furiously across the raging river. \item A rustic boat bobs gently on the calm and tranquil river.
\end{itemize}\\

\rowcolor{rowx} breakdance-flare & a man dances on the road. &
\begin{itemize}[topsep=0pt]
\item A young man elegantly dances on the deserted road under the starry night sky. \item The handsome man dances enthusiastically on the bumpy dirt road, kicking up dust as he moves. \item A man gracefully dances on the winding road, surrounded by the picturesque mountain scenery. \item The athletic man dances energetically on the long and straight road, his sweat glistening under the bright sun. \item The talented man dances flawlessly on the busy city road, attracting a crowd of mesmerized onlookers.
\end{itemize}\\

bus & a bus moves on the street. &
\begin{itemize}[topsep=0pt]
\item A big red bus swiftly maneuvers through the crowded city streets. \item A sleek silver bus gracefully glides down the busy urban avenue. \item A colorful double-decker bus boldly navigates through the bustling downtown district. \item A vintage yellow bus leisurely rolls down the narrow suburban road. \item A modern electric bus silently travels along the winding coastal highway.
\end{itemize}\\

\rowcolor{rowx} camel & a camel walks on the desert. &
\begin{itemize}[topsep=0pt]
\item A majestic camel gracefully strides across the scorching desert sands. \item A lone camel strolls leisurely through the vast, arid expanse of the desert. \item A humpbacked camel plods methodically across the barren and unforgiving desert terrain. \item A magnificent camel marches stoically through the seemingly endless desert wilderness. \item A weathered camel saunters across the sun-scorched dunes of the desert, its gaze fixed on the horizon.
\end{itemize}\\

 car-roundabout & a jeep turns on a road. &
\begin{itemize}[topsep=0pt]
\item A shiny red jeep smoothly turns on a narrow, winding road in the mountains. \item A rusty old jeep suddenly turns on a bumpy, unpaved road in the countryside. \item A sleek black jeep swiftly turns on a deserted, dusty road in the desert. \item A modified green jeep expertly turns on a steep, rocky road in the forest. \item A gigantic yellow jeep slowly turns on a wide, smooth road in the city.
\end{itemize}\\

\rowcolor{rowx} car-shadow & a car moves to a building. &
\begin{itemize}[topsep=0pt]
\item A sleek black car swiftly glides towards a towering skyscraper. \item A shiny silver vehicle gracefully maneuvers towards a modern glass building. \item A vintage red car leisurely drives towards an abandoned brick edifice. \item A luxurious white car elegantly approaches a stately colonial mansion. \item A rusty blue car slowly crawls towards a dilapidated concrete structure.
\end{itemize}\\

 car-turn & a jeep on a forest road. &
\begin{itemize}[topsep=0pt]
\item A shiny silver jeep was maneuvering through the dense forest, kicking up dirt and leaves in its wake. \item A dusty old jeep was making its way down the winding forest road, creaking and groaning with each bump and turn. \item A sleek black jeep was speeding along the narrow forest road, dodging trees and rocks with ease. \item A massive green jeep was lumbering down the rugged forest road, its powerful engine growling as it tackled the steep incline. \item A rusty red jeep was bouncing along the bumpy forest road, its tires kicking up mud and gravel as it went.
\end{itemize}\\

\rowcolor{rowx} cows & a cow walks on the grass. &
\begin{itemize}[topsep=0pt]
\item A spotted cow leisurely grazes on the lush, emerald-green grass. \item A contented cow ambles across the dewy, verdant pasture. \item A brown cow serenely strolls through the sun-kissed, rolling hills. \item A beautiful cow saunters through the vibrant, blooming meadow.\item A gentle cow leisurely walks on the soft, velvety green grass, enjoying the warm sunshine.
\end{itemize}\\

\cmidrule[0.75pt]{1-3}
\end{tabular}}
\label{tab:prompts_control}
\end{table*}

\setcounter{table}{1}

\begin{table*}[h!] %
\centering
\caption{\textbf{Sequence-prompt pairs used to evaluate ControlNet-based pipelines}: All sequences are from DAVIS \cite{davis2017}. These pairs correspond to the original ControlVideo \cite{controlvideo} quantitative evaluation setting. [continued...]}
\tablestyle{1.8pt}{1.}
\resizebox{0.9\textwidth}{!}{
\scriptsize
\begin{tabular}{m{0.14\textwidth} m{0.20\textwidth} m{0.78\textwidth}}
\cmidrule[0.75pt]{1-3}
\multirow{1}{*}{Sequence} & Source prompt & $\;\;\;$Target prompts \\
\cmidrule{1-3}

\rowcolor{rowx} dog & a dog walks on the ground. &
\begin{itemize}[topsep=0pt]
\item A fluffy brown dog leisurely strolls on the grassy field. \item A scruffy little dog energetically trots on the sandy beach. \item A majestic black dog gracefully paces on the polished marble floor. \item A playful spotted dog joyfully skips on the leaf-covered path. \item A curious golden dog curiously wanders on the rocky mountain trail.
\end{itemize}\\

 elephant & an elephant walks on the ground. &
\begin{itemize}[topsep=0pt]
\item A massive elephant strides gracefully across the dusty savannah. \item A majestic elephant strolls leisurely along the lush green fields. \item A mighty elephant marches steadily through the rugged terrain. \item A gentle elephant ambles peacefully through the tranquil forest. \item A regal elephant parades elegantly down the bustling city street.
\end{itemize}\\

\rowcolor{rowx} flamingo & a flamingo wanders in the water. &
\begin{itemize}[topsep=0pt]
\item A graceful pink flamingo leisurely wanders in the cool and refreshing water, its slender legs elegantly stepping on the soft sand. \item A vibrant flamingo casually wanders in the clear and sparkling water, its majestic wings spread wide in the sunshine. \item A charming flamingo gracefully wanders in the calm and serene water, its delicate neck curving into an elegant shape. \item A stunning flamingo leisurely wanders in the turquoise and tranquil water, its radiant pink feathers reflecting the shimmering light. \item A magnificent flamingo elegantly wanders in the sparkling and crystal-clear water, its striking plumage shining brightly in the sun.
\end{itemize}\\

 gold-fish & golden fishers swim in the water. & 
\begin{itemize}[topsep=0pt]
\item Majestic golden fishers glide gracefully in the crystal-clear waters. \item Brilliant golden fishers swim serenely in the shimmering blue depths. \item Glittering golden fishers dance playfully in the glistening aquamarine waves. \item Gleaming golden fishers float leisurely in the peaceful turquoise pools. \item Radiant golden fishers meander lazily in the tranquil emerald streams.
\end{itemize}\\

\rowcolor{rowx} hike & a man hikes on a mountain. &
\begin{itemize}[topsep=0pt]
\item A rugged man is trekking up a steep and rocky mountain trail. \item A fit man is leisurely hiking through a lush and verdant forest. \item A daring man is scaling a treacherous and jagged peak in the alpine wilderness. \item A seasoned man is exploring a remote and rugged canyon deep in the desert. \item A determined man is trudging up a snowy and icy mountain slope, braving the biting cold and fierce winds.
\end{itemize}\\

hockey & a player is playing hockey on the ground. &
\begin{itemize}[topsep=0pt]
\item A skilled player is furiously playing ice hockey on the smooth, glistening rink. \item A young, agile player is energetically playing field hockey on the lush, green grass. \item An experienced player is gracefully playing roller hockey on the sleek, polished pavement. \item A determined player is passionately playing street hockey on the gritty, urban asphalt. \item A talented player is confidently playing air hockey on the fast-paced, neon-lit table.
\end{itemize}\\

\rowcolor{rowx} kite-surf & a man is surfing on the sea. &
\begin{itemize}[topsep=0pt]
\item A muscular man is expertly surfing the gigantic waves of the Pacific Ocean. \item A handsome man is gracefully surfing on the crystal clear waters of the Caribbean Sea. \item A daring man is fearlessly surfing through the dangerous, choppy waters of the Atlantic Ocean. \item An athletic man is skillfully surfing on the wild and untamed waves of the Indian Ocean. \item A young man is confidently surfing on the smooth, peaceful waters of a serene lake.
\end{itemize}\\

lab-coat & three women stands on the lawn. &
\begin{itemize}[topsep=0pt]
\item Three stunning women are standing elegantly on the lush green lawn, chatting and laughing. \item Three young and vibrant women are standing proudly on the well-manicured lawn, enjoying the sunshine. \item Three fashionable women in colorful dresses are standing gracefully on the emerald green lawn, taking selfies. \item Three confident women with radiant smiles are standing tall on the soft, green lawn, enjoying the fresh air. \item Three beautiful women, each dressed in their own unique style, are standing on the lush and verdant lawn, admiring the scenery.
\end{itemize}\\

\cmidrule[0.75pt]{1-3}
\end{tabular}}
\label{tab:prompts_control}
\end{table*}

\setcounter{table}{1}

\begin{table*}[h!] %
\centering
\caption{\textbf{Sequence-prompt pairs used to evaluate ControlNet-based pipelines}: All sequences are from DAVIS \cite{davis2017}. These pairs correspond to the original ControlVideo \cite{controlvideo} quantitative evaluation setting.}
\tablestyle{1.8pt}{1.}
\resizebox{0.9\textwidth}{!}{
\scriptsize
\begin{tabular}{m{0.14\textwidth} m{0.20\textwidth} m{0.78\textwidth}}
\cmidrule[0.75pt]{1-3}
\multirow{1}{*}{Sequence} & Source prompt & $\;\;\;$Target prompts \\
\cmidrule{1-3}

\rowcolor{rowx} longboard & a man is playing skateboard on the alley. &
\begin{itemize}[topsep=0pt]
\item A young man is skillfully skateboarding on the busy city street, weaving in and out of the crowds with ease. \item An experienced skateboarder is fearlessly gliding down a steep, curvy road on his board, executing impressive tricks along the way. \item A daring skater is performing gravity-defying flips and spins on his board, effortlessly navigating through a challenging skatepark course. \item A talented skateboarder is carving up the smooth pavement of an empty parking lot, creating beautiful patterns with his board and body. \item A passionate skater is practicing his moves on a quiet neighborhood street, with the sound of his board echoing through the peaceful surroundings.
\end{itemize}\\

 mallard-water & a mallard swims on the water. &
\begin{itemize}[topsep=0pt]
\item A vibrant mallard glides gracefully on the shimmering water. \item A beautiful mallard paddles through the calm, blue water. \item A majestic mallard swims elegantly on the tranquil lake. \item A striking mallard floats effortlessly on the sparkling pond. \item A colorful mallard glides smoothly on the rippling surface of the water.
\end{itemize}\\

\rowcolor{rowx} mbike-trick & a man riding motorbike. &
\begin{itemize}[topsep=0pt]
\item A young man riding a sleek, black motorbike through the winding mountain roads. \item An experienced man effortlessly riding a powerful, red motorbike on the open highway. \item A daring man performing gravity-defying stunts on a high-speed, blue motorbike in an empty parking lot. \item A confident man cruising on a vintage, yellow motorbike along the picturesque coastal roads. \item A rugged man maneuvering a heavy, dusty motorbike through the rugged terrain of a desert.
\end{itemize}\\

\rowcolor{rowx} rhino & a rhino walks on the rocks. &
\begin{itemize}[topsep=0pt]
\item A massive rhino strides confidently across the jagged rocks. \item A majestic rhino gracefully navigates the rugged terrain of the rocky landscape. \item A powerful rhino marches steadily over the rough and rocky ground. \item A colossal rhino plods steadily through the craggy rocks, undeterred by the challenging terrain. \item A sturdy rhino confidently traverses the treacherous rocks with ease.
\end{itemize}\\

 surf & a sailing boat moves on the sea. &
\begin{itemize}[topsep=0pt]
\item A graceful sailing boat glides smoothly over the tranquil sea. \item A sleek sailing boat cuts through the shimmering sea with ease. \item A majestic sailing boat cruises along the vast, azure sea. \item A vintage sailing boat bobs gently on the calm, turquoise sea. \item A speedy sailing boat races across the glistening, open sea.
\end{itemize}\\

\rowcolor{rowx} swing & a girl is playing on the swings. &
\begin{itemize}[topsep=0pt]
\item A young girl with pigtails is joyfully swinging on the colorful swings in the playground. \item The little girl, giggling uncontrollably, is happily playing on the old-fashioned wooden swings. \item A blonde girl with a big smile on her face is energetically playing on the swings in the park. \item The girl, wearing a flowery dress, is gracefully swaying back and forth on the swings, enjoying the warm breeze. \item A cute little girl, dressed in a red coat, is playfully swinging on the swings, her hair flying in the wind.
\end{itemize}\\

 tennis & a man is playing tennis. &
\begin{itemize}[topsep=0pt]
\item The skilled man is effortlessly playing tennis on the court. \item A focused man is gracefully playing a game of tennis. \item A fit and agile man is playing tennis with precision and finesse. \item A competitive man is relentlessly playing tennis with his opponent. \item The enthusiastic man is eagerly playing a game of tennis, sweat pouring down his face.
\end{itemize}\\
	
\rowcolor{rowx} walking & a selfie of walking man. &
\begin{itemize}[topsep=0pt]
\item A stylish young man takes a selfie while strutting confidently down the busy city street. \item An energetic man captures a selfie mid-walk, showcasing his adventurous spirit. \item A happy-go-lucky man snaps a selfie as he leisurely strolls through the park, enjoying the sunny day. \item A determined man takes a selfie while briskly walking towards his destination, never breaking stride. \item A carefree man captures a selfie while wandering aimlessly through the vibrant cityscape, taking in all the sights and sounds.
\end{itemize}\\

\cmidrule[0.75pt]{1-3}
\end{tabular}}
\label{tab:prompts_control}
\end{table*}

%% file: texts/supplementary/B_results.tex
\renewcommand{\thetable}{B.\arabic{table}}
\renewcommand{\thefigure}{B.\arabic{figure}}
\renewcommand{\thesection}{B}
\renewcommand{\thesubsection}{B.\arabic{subsection}}
\setcounter{table}{0}
\setcounter{figure}{0}

\newpage
\section{Additional results}
\label{sec:sup_B}
\subsection{Qualitative comparisons}
We show additional qualitative comparisons for inversion-based pipelines in \fref{fig:fatezero_temporal_consistency}. Here, we mainly focus on shape editing, and present multiple edited frames of each sequence using FateZero \cite{fatezero}, Tune-A-Video \cite{tuneavideo}, TokenFlow \cite{tokenflow} and Frame SDEdit \cite{sdedit}. Tune-A-Video requires 1-shot finetuning on a given sequence, whereas TokenFlow and Frame SDEdit are based on stable diffusion \cite{stablediffusion} checkpoints. FateZero and our implementation rely on Tune-A-Video checkpoints for shape editing, without needing any further finetuning for their respective proposed improvements. Frame SDEdit shows no consistency among frames, being an image editing pipeline. Among video editing pipelines, ours show the best fidelity and temporal-consistency, while also generating outputs faster (see latency measurements given in Table 1 and Fig.~5 in the main paper). Notably, thanks to Object-Centric Sampling, our pipeline gives more-faithful background reconstructions, as such regions are expected to be un-edited based on the given shape editing prompts.

In \fref{fig:controlvideo_sup_temporal_consistency}, we show additional qualitative comparisons for ControlNet-based pipelines sucha as ControlVideo \cite{controlvideo} and Text2Video-Zero \cite{text2video-zero}. Here, all methods are conditioned on  Depth-maps, while using SD \cite{stablediffusion} checkpoints without further finetuning. OCD shows comparable performance with its baseline ControlVideo, while being significantly faster (see latency measurements given in Table 2 in the main paper). It is also more temporally-consistent compared to Text2Video-Zero which uses sparse instead of dense cross-frame attention, while having comparable latency.
\clearpage
\input{figures/fatezero_temporal_consistency}
\clearpage
\input{figures/controlvideo_temporal_consistency}
\clearpage
\subsection{Quantitative comparisons}
\paragraph{Other baselines for latency reduction} We discuss simple baselines for reducing latency in \tref{tab:sota_fatezero_supp} and \tref{tab:sota_controlvideo_supp}. We report both fidelity (Temporal-consistency, CLIP-score) and latency (inversion, generation, UNet \cite{unet} time). Here, UNet time corresponds to just running UNet inference without any additional overheads (\eg memory access), which we use to highlight the cost of such overheads. For ControlVideo \cite{controlvideo} baselines, we show results with either Depth or Canny-edge conditioning. 

In both inversion-based and ControlNet-based settings, we devise our optimized-baselines by reducing diffusion steps 50$\rightarrow$20 and applying token merging, which give reasonable reconstructions. This is our default starting point for implementing OCD. Going further, we also consider diff.~steps=5, which fails to retain details in reconstructions. Instead of token merging, we can also apply pooling strategies on key-value tokens. Despite giving similar speed-ups, these result in sub-par performance compared to ToMe, especially in shape editing (although not always captured in quantitative numbers). In ControlVideo setup, we can choose to do merging on both UNet  and ControlNet \cite{controlnet} models, resulting in further speed-ups with a minimal drop in fidelity. We further observe that we can re-use the same control signal for multiple diffusion steps, allowing us to run ControlNet at a reduced rate (Reduced/Single inference in \tref{tab:sota_controlvideo_supp}).
\input{tables/supp_metrics_fatezero}

\input{tables/supp_metrics_controlvideo}
\input{tables/supp_storage_cost}

\paragraph{Cost of memory access vs.~computations}
Inversion-based editing pipelines rely on guidance from the inversion process during generation (\eg based on latents~\cite{plugandplay} or attention maps~\cite{fatezero}). 
When running inference, such features need to be stored (which may include additional GPU$\rightarrow$RAM transfer), accessed and reused. 
This cost can be considerable, especially for resource-constrained hardware. This cost measured in latency, is shown in \tref{tab:sota_fatezero_supp}, as the difference between inversion and UNet times. Alternatively, it can also be seen as the storage requirement as given in \tref{tab:storage_cost}.
On FateZero~\cite{fatezero}, we observe that the storage cost is indeed significant, and affects the latency more than the computations.
With OCD, we directly reduce the cost for attention computation, storage and access.
\input{tables/supp_ablation_fatezero_blending_controlled}

\paragraph{Expected savings of Object-Centric Sampling} We run a control experiment to observe the expected latency reductions when using our Object-Centric Sampling, at different object sizes ($\Delta$) and Blending step ratios ($\gamma$), given in \tref{tab:ablation_blending_controlled}. Here, we consider hypothetical inputs, so that we can ablate the settings more-clearly. The baseline sees inputs of size $\Delta=64\times 64$, and runs all diffusion steps at full-resolution ($\gamma=1.0$). In our Object-Centric Sampling, we use $\gamma=0.25$ by default, whereas $\Delta$ depends on objects in a given sequence. As expected, we can obtain the most savings with fewer blending steps and when foreground objects are smaller in size. A user may refer to this guide to get an idea about the expected savings.

\input{figures/rebuttal_survey}
\paragraph{User preference study} We conduct a user study to measure the editing quality of OCD w.r.t.~(1)~Optimized-FateZero and (2)~Blended Latent Diffusion \cite{avrahami2023blended}.
The study consists of randomized A/B preferences tests, where observers were asked to assess video edits in overall quality as well as temporal consistency and alignment with edit prompt.
Based on 1143 responses collected from 37 participants, we find that OCD is likely preferred by users in all assessments.
More specifically, this study testifies the benefit of OCD w.r.t. the optimized baseline beyond computational savings, as its edits are preferred 77\% of the time, and rewarded by users in temporal consistency (75\%) and prompt alignment (66\%).

\paragraph{Other methods with disentangled diffusion sampling}
Both OCD and Blended Latent Diffusion (BLD) \cite{avrahami2023blended} use saliency masks to disentangle foreground and background during diffusion sampling.
Although BLD focuses on local edits, its design does not imply any efficiency gains (as it processes background, but discards during blending).
On the contrary, our proposal trades-off computational cost in the background, allowing for a better foreground edit at a reduced latency. 
Such a change in scope results in several key differences.
In OCD, background and foreground latents undergo two \emph{separate} (even, parallelizable) sampling processes, operating at different resolutions and sampling rates, before being blended at a certain pre-defined step. 
Differently, latents in BLD are blended at every step of the \emph{same} diffusion process, and its latency is by design lower bounded by that of a standard diffusion process. 
To compare it with our approach, we included BLD edits in the user study in \fref{fig:user_study}, where we observe OCD is preferred most of the times in terms of temporal consistency (65.35\%) and prompt alignment (60.71\%) while also being faster (17.51s vs 19.68s). In metrics, OCD is comparable to BLD both in temporal consistency (0.967 vs 0.968) and clip score (0.331 vs 0.329) respectively.

%% file: figures/fatezero_temporal_consistency.tex
\begin{figure*}[h!]
\centering
\resizebox{0.95\textwidth}{!}{
\begin{tabular}{cc}
\hspace{5mm}\includegraphics[height=20cm]{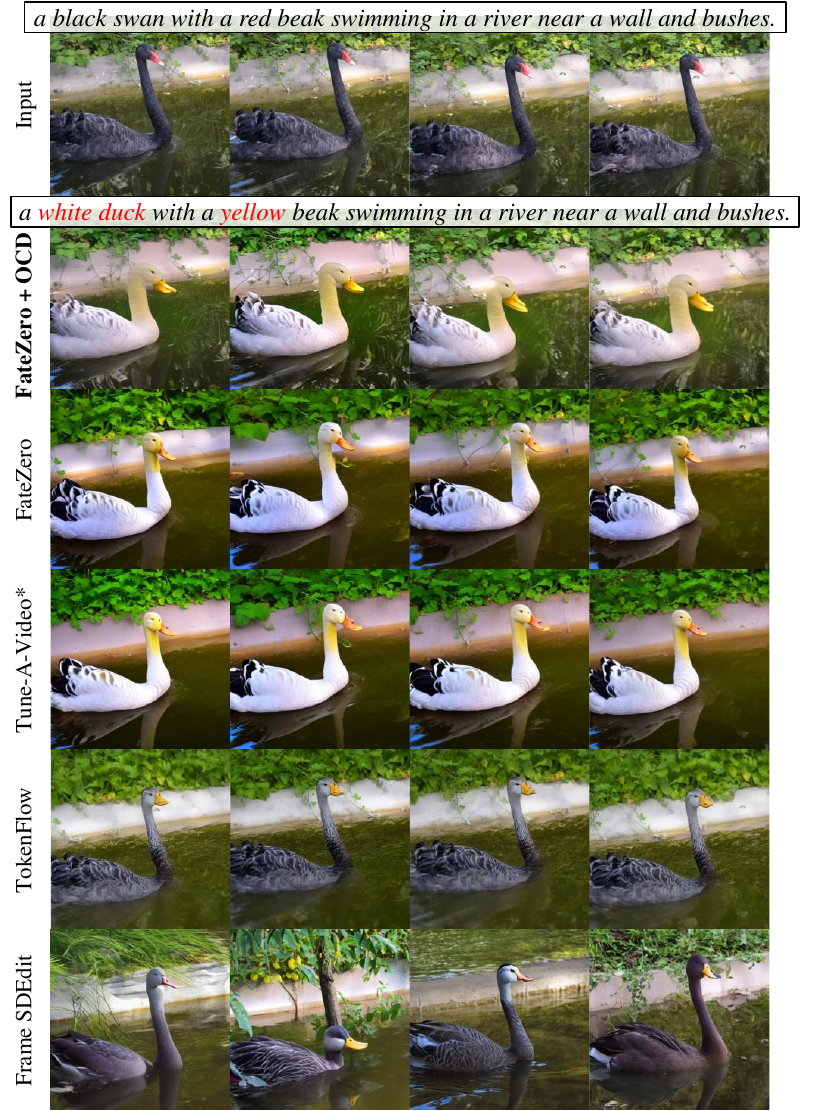}&
\hspace{-8mm}\includegraphics[height=20cm]{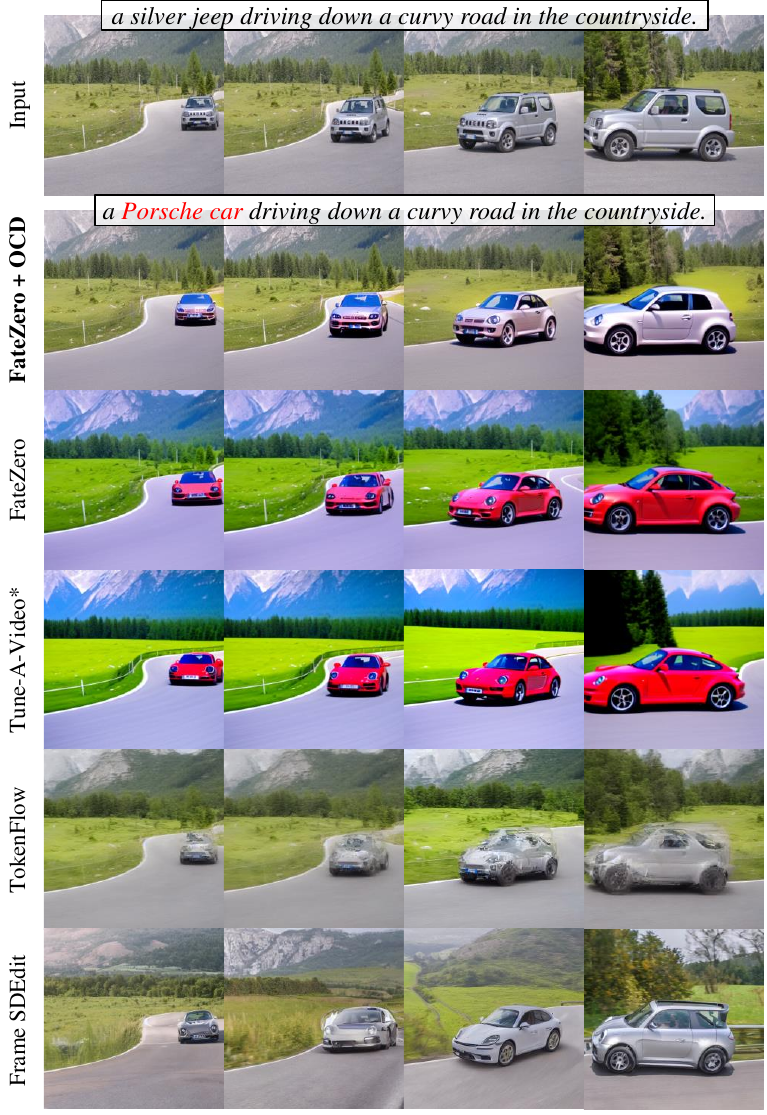}\\
\includegraphics[height=20cm]{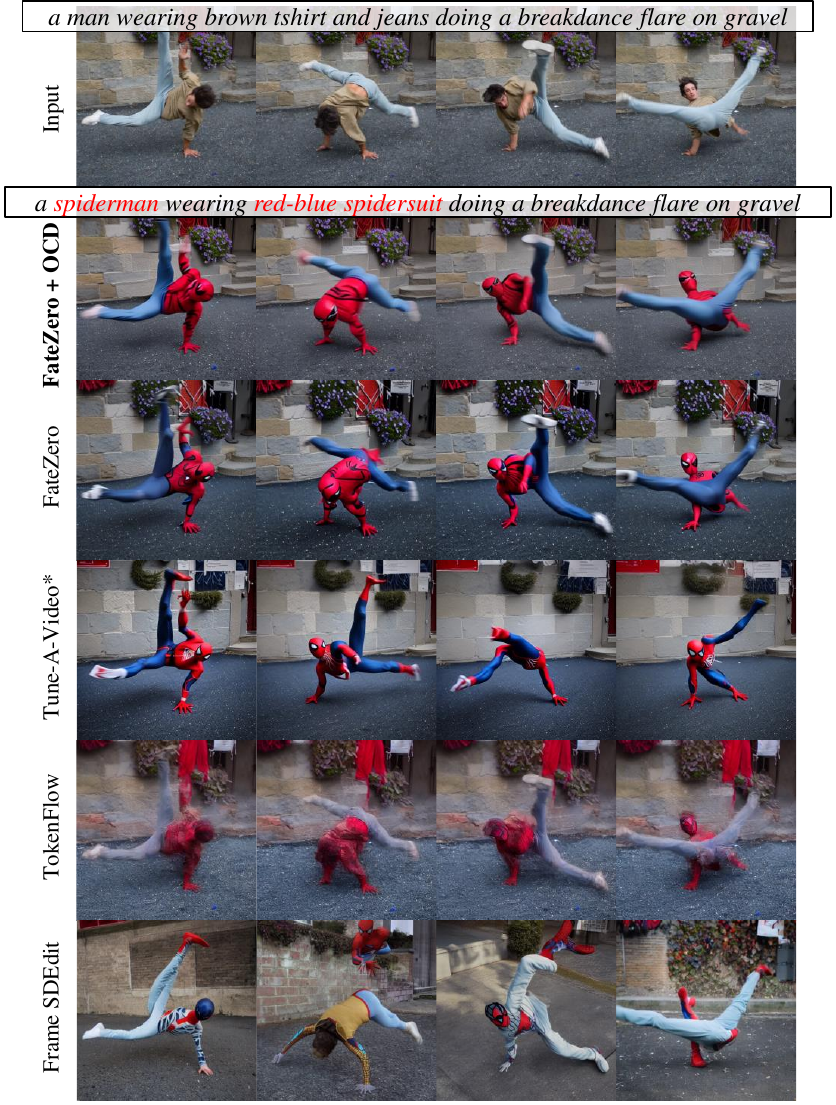}&
\includegraphics[height=20cm]{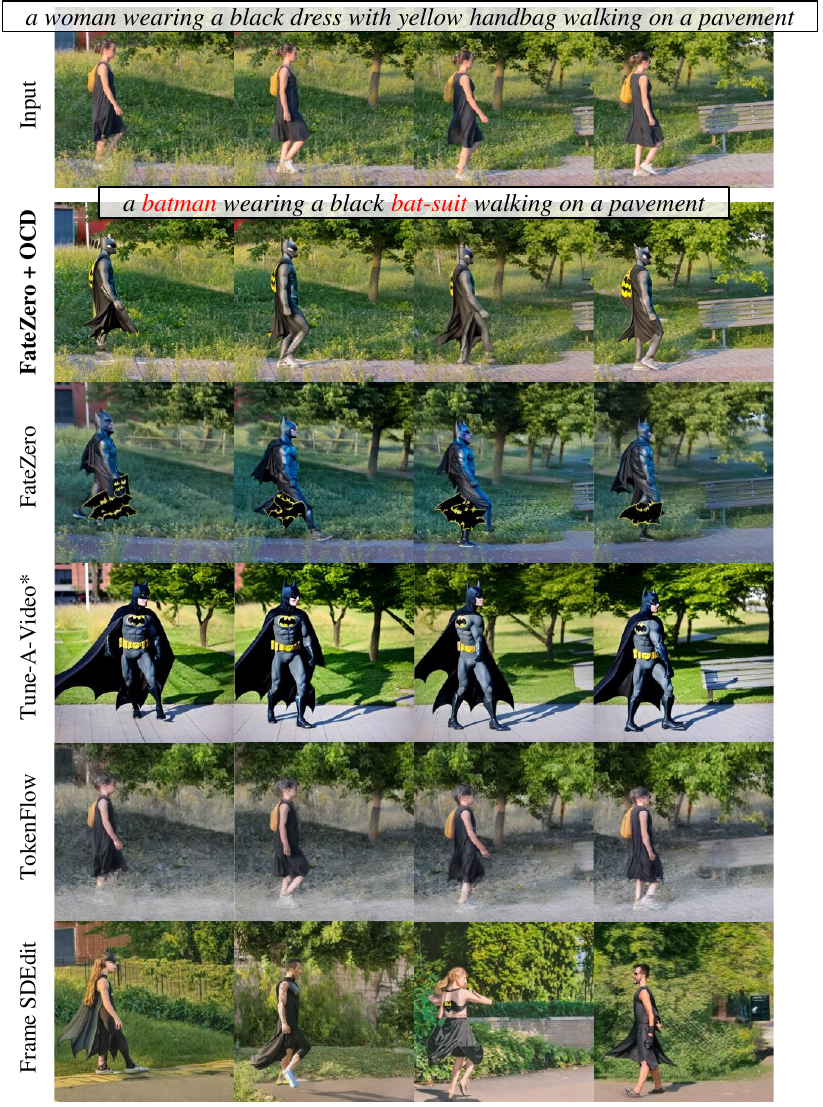}\\
\end{tabular}}
\caption{\textbf{Qualitative comparison on \emph{blackswan}, \emph{car-turn}, \emph{breakdance-flare} and \emph{lucia} sequences \cite{davis2017}:} We show shape editing results of our method (Optimized-FateZero + OCD), in comparison with FateZero \cite{fatezero}, Tune-A-Video \cite{tuneavideo}, TokenFlow \cite{tokenflow} and SDEdit \cite{sdedit}. Our results show better semantic quality (\eg alignement with target prompt) and visual fidelity (\eg temporal consistency, faithful background), while also being more efficient (Table~1 in the main paper). Best viewed zoomed-in.}
\label{fig:fatezero_temporal_consistency}
\end{figure*}

%% file: figures/controlvideo_temporal_consistency.tex
\begin{figure}[h!]
\centering
\resizebox{0.7\textwidth}{!}{
\begin{tabular}{c}
\hspace{-5.5mm}
\includegraphics[height=19.15cm]{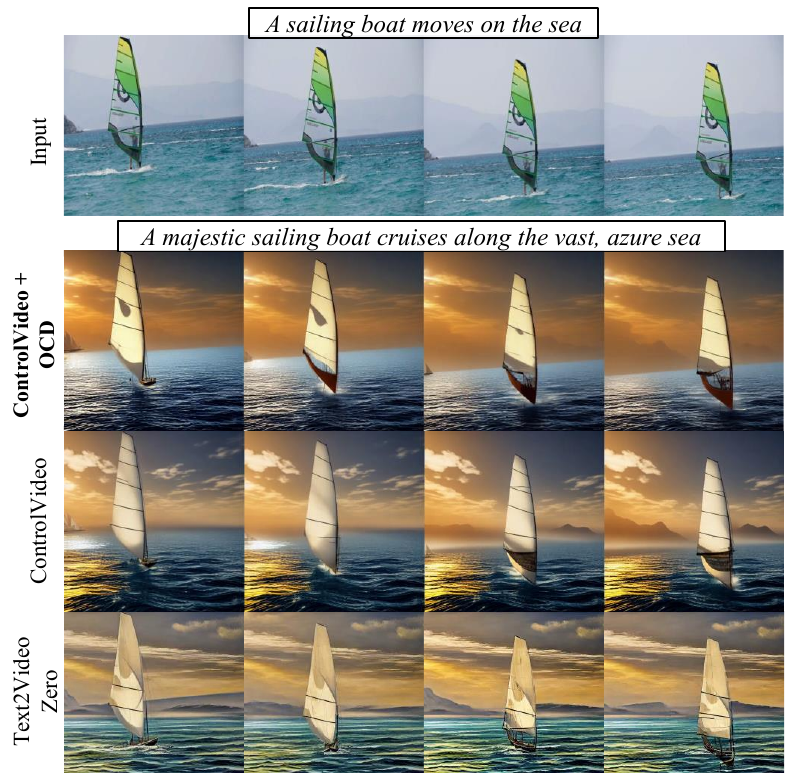}\\
\includegraphics[height=20cm]{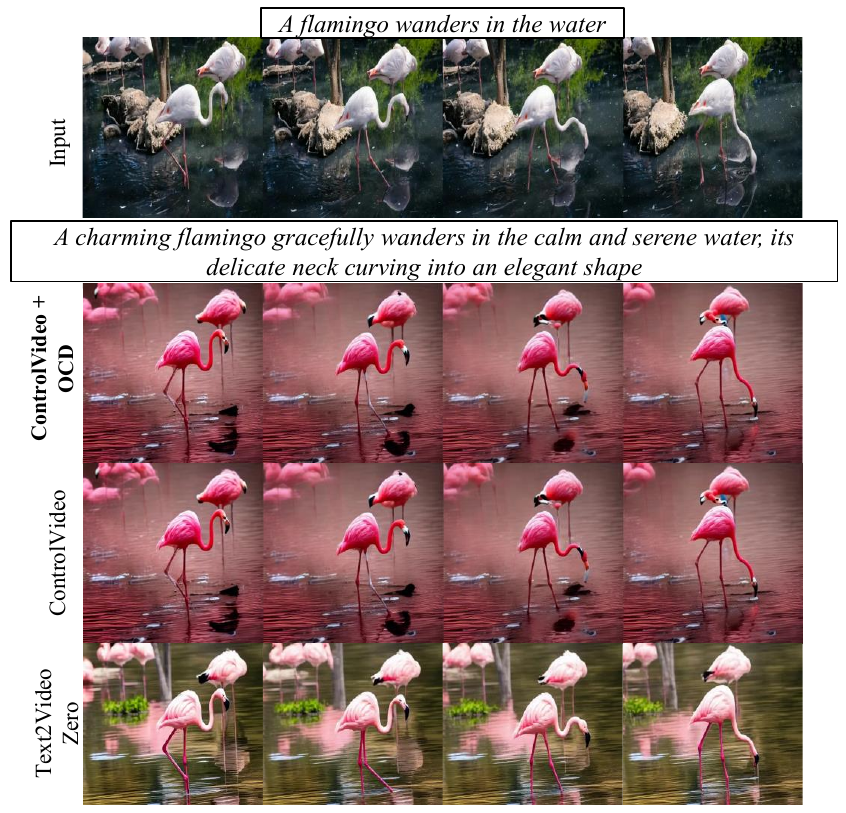}
\end{tabular}}
\caption{\textbf{Qualitative comparison on \emph{surf} and \emph{flamingo} sequences \cite{davis2017}:} We show shape editing results of our method (Optimized-ControlVideo + OCD), in comparison with ControlVideo \cite{controlvideo} and Text2Video-Zero \cite{text2video-zero}. All methods use Depth conditioning. Our results show comparable quality with baseline ControlVideo while being significantly faster (Tab~2 in the main paper), and better temporal consistency compared to Text2Video-Zero. Best viewed zoomed-in.}
\label{fig:controlvideo_sup_temporal_consistency}
\end{figure}

%% file: tables/supp_metrics_fatezero.tex
\begin{table}[t!] %
\centering
\caption{\textbf{Additional FateZero \cite{fatezero} baselines:} We report CLIP metrics of fidelity (Temporal-consistency, CLIP-score) and latency (inversion, generation, UNet \cite{unet} time). The difference between inversion and UNet time corresponds to other overheads, dominated by memory access. Fewer diffusion steps (\eg 5) and pooling operations can also gain significant speed-ups, but break reconstructions (not always visible in fidelity metrics).}
\tablestyle{1.8pt}{1.}
\resizebox{0.7\columnwidth}{!}{
\begin{tabular}{lccrrr}
\toprule
\multirow{2}{*}{Model} & \multicolumn{2}{c}{CLIP metrics $\uparrow$} & \multicolumn{3}{c}{Latency (s) $\downarrow$}\\
\cmidrule( r{0.5mm}){2-3} \cmidrule(l{0.5mm}){4-6}
& Tem-con & Cl-score & \multicolumn{1}{c}{Inv} & \multicolumn{1}{c}{Gen} & \multicolumn{1}{c}{UNet}\\ %
\midrule
FateZero~\cite{fatezero} & 0.961 & 0.344 & 135.80 & 41.34 & 20.63 \\ %
\quad $+$ diff. steps=5 & 0.968 & 0.306 & 14.84 & 4.98 & 2.17 \\ %
\quad $+$ diff. steps=20 & 0.961 & 0.341 & 61.82 & 18.41 & 8.03 \\ %
\qquad $+$ avg. pool (4,4) & 0.958 & 0.335 & 9.91 & 11.80 & 7.08 \\ %
\qquad $+$ max pool (4,4) & 0.959 & 0.275 & 9.96 & 11.91 & 6.91 \\ %
\rowcolor{row} Optimzed-FateZero & 0.966 & 0.334 & 9.54 & 10.14 & 7.79 \\ %
\rowcolor{row} \quad $+$ OCD & 0.967 & 0.331 & 8.22 & 9.29 & 6.38 \\ %
\bottomrule
\end{tabular}}
\label{tab:sota_fatezero_supp}
\end{table}

%% file: tables/supp_metrics_controlvideo.tex
\begin{table}[t!] %
\centering
\caption{\textbf{Additional ControlVideo \cite{controlvideo} baselines:} We report CLIP metrics of fidelity (Temporal-consistency, CLIP-score) with either Depth or \textcolor{gray}{(Canny-edge)} conditioning, and latency (generation, UNet \cite{unet} time). The difference between generation and UNet time corresponds to other overheads, dominated by ControlNet \cite{controlnet}. Fewer diffusion steps (\eg 5) and pooling operations can also gain significant speed-ups, but break reconstructions (not always visible in fidelity metrics). We also observe that ControlNet inference need not be done at the same frequency as denoising, which can lead to further speed-ups. 
}
\tablestyle{1.8pt}{1.}
\resizebox{0.8\columnwidth}{!}{
\begin{tabular}{lccrr}
\toprule
\multirow{2}{*}{Model} & \multicolumn{2}{c}{CLIP metrics $\uparrow$} & \multicolumn{2}{c}{Latency (s) $\downarrow$}\\
\cmidrule(r{0.5mm}){2-3} \cmidrule(l{0.5mm}){4-5}
& Tem-con & Cl-score & \multicolumn{1}{c}{Gen} & \multicolumn{1}{c}{UNet}\\ 
\midrule
ControlVideo~\cite{controlvideo} & 0.972 \textcolor{gray}{(0.968)} & 0.318 \textcolor{gray}{(0.308)} & 152.64 & 137.68 \\
\quad $+$ diff. steps=5 & 0.978 \textcolor{gray}{(0.971)} & 0.309 \textcolor{gray}{(0.295)} & 19.58 & 13.58 \\
\quad $+$ diff. steps=20 & 0.978 \textcolor{gray}{(0.971)} & 0.316 \textcolor{gray}{(0.304)} & 64.61 & 54.83 \\
\qquad $+$ avg.pool (2,2) & 0.977 \textcolor{gray}{(0.968)} & 0.309 \textcolor{gray}{(0.295)} & 30.53 & 20.56 \\
\qquad $+$ max.pool (2,2) & 0.972 \textcolor{gray}{(0.973)} & 0.225 \textcolor{gray}{(0.212)} & 30.32 & 20.53 \\
\rowcolor{row} Optimized-ControlVideo & 0.978 \textcolor{gray}{(0.972)} & 0.314 \textcolor{gray}{(0.303)} & 31.12 & 21.42 \\
\rowcolor{row} \quad $+$ OCD & 0.977 \textcolor{gray}{(0.967)} 
& 0.313 \textcolor{gray}{(0.302)} & 25.21 & 15.61 \\
\rowcolor{row} \quad $+$ OCD (UNet, ControlNet) & 0.976 \textcolor{gray}{(0.969)} & 0.306 \textcolor{gray}{(0.297)} & 25.13 & 15.41 \\
\rowcolor{row} \qquad $+$ ControlNet Red. Inf. & 0.977 \textcolor{gray}{(0.968)} & 0.313 \textcolor{gray}{(0.301)} & 23.62 & 15.47 \\
\rowcolor{row} \qquad $+$ ControlNet Sin. Inf. & 0.973 \textcolor{gray}{(0.964)} & 0.307 \textcolor{gray}{(0.293)} & 22.35 & 15.48 \\
\bottomrule
\end{tabular}}
\label{tab:sota_controlvideo_supp}
\end{table}

%% file: tables/supp_storage_cost.tex
\begin{table}[t!] %
\centering
\caption{\textbf{Memory requirement for attention maps}: In FateZero \cite{fatezero} setting, we show additional baselines and the corresponding storage requirements which directly affect the memory-access overhead.
FateZero stores attention maps of all UNet \cite{unet} blocks for all diffusion steps. Our contributions help reduce this cost. It can potentially enable attention maps to be kept on GPU memory itself (w/o having to move between GPU and RAM), further improving latency. Each float is stored in 16bits.}
\tablestyle{1.8pt}{1.}
\resizebox{0.45\columnwidth}{!}{
\begin{tabular}{lr}
\toprule
\multirow{1}{*}{Model} & Disk-space (GB) $\downarrow$\\
\midrule
FateZero & 74.54\\
\quad $+$ diff. steps=5 & 7.45\\
\quad $+$ diff. steps=20 & 29.82\\
\qquad $+$ pool (4,4) & 3.06\\
\rowcolor{row} Optimized-FateZero & 5.05\\
\rowcolor{row} \quad $+$ OCD & 4.22\\
\bottomrule
\end{tabular}}
\label{tab:storage_cost}
\end{table}

%% file: tables/supp_ablation_fatezero_blending_controlled.tex
\begin{table}[t!] %
\centering
\caption{\textbf{Control experiment on Object-Centric Sampling:} We evaluate the latency savings at different \textit{hypothetical} object sizes ($\Delta$) and blending step ratios ($\gamma$). The baseline is with $\Delta=64\times64$ and $\gamma=1.0$ (with total 20 diffusion steps). We can get the most savings at a smaller object size and blending step ratio. It is worth noting that this control experiment does not correspond to actual sequence-prompt pairs, and is just intended to give the reader an idea about expected savings.}
\tablestyle{1.8pt}{1.}
\resizebox{0.7\columnwidth}{!}{
\begin{tabular}{l ccc ccc ccc cc}
\toprule
\multicolumn{1}{l}{\multirow{2.5}{*}{Blending steps}} & \multicolumn{11}{c}{Latency (s) @ \#tokens ($\Delta$) $\downarrow$} \\
\cmidrule(l{1mm}){2-12}
\multicolumn{1}{l}{\multirow{-0.8}{*}{ratio ($\gamma$)}} & \multicolumn{2}{c}{$64\times64$} && \multicolumn{2}{c}{$48\times48$} && \multicolumn{2}{c}{$32\times32$} && \multicolumn{2}{c}{$16\times16$} \\
\cmidrule(l{0.5mm}r{0.3mm}){2-3} \cmidrule{5-6} \cmidrule{8-9} \cmidrule{11-12}
& \multicolumn{1}{c}{Inv} & \multicolumn{1}{c}{Gen} && \multicolumn{1}{c}{Inv} & \multicolumn{1}{c}{Gen} && \multicolumn{1}{c}{Inv} & \multicolumn{1}{c}{Gen} && \multicolumn{1}{c}{Inv} & \multicolumn{1}{c}{Gen} \\
\midrule
1.00 & 12.31 & 10.39 && - & - && - & - && - & \multicolumn{1}{c}{-} \\
0.50 & - & - && 9.56 & 8.67 && 9.05 & 7.60 && 8.25 & 7.01 \\
0.25 & - & - && 8.03 & 7.94 && 7.11 & 6.00 && 5.90 & 4.96 \\
0.05 & - & - && 6.82 & 7.17 && 5.85 & 4.99 && 4.43 & 3.80 \\
\bottomrule
\end{tabular}}
\label{tab:ablation_blending_controlled}
\end{table}

%% file: figures/rebuttal_survey.tex
\begin{figure}[t!]
\centering
\includegraphics[width=0.85\linewidth]{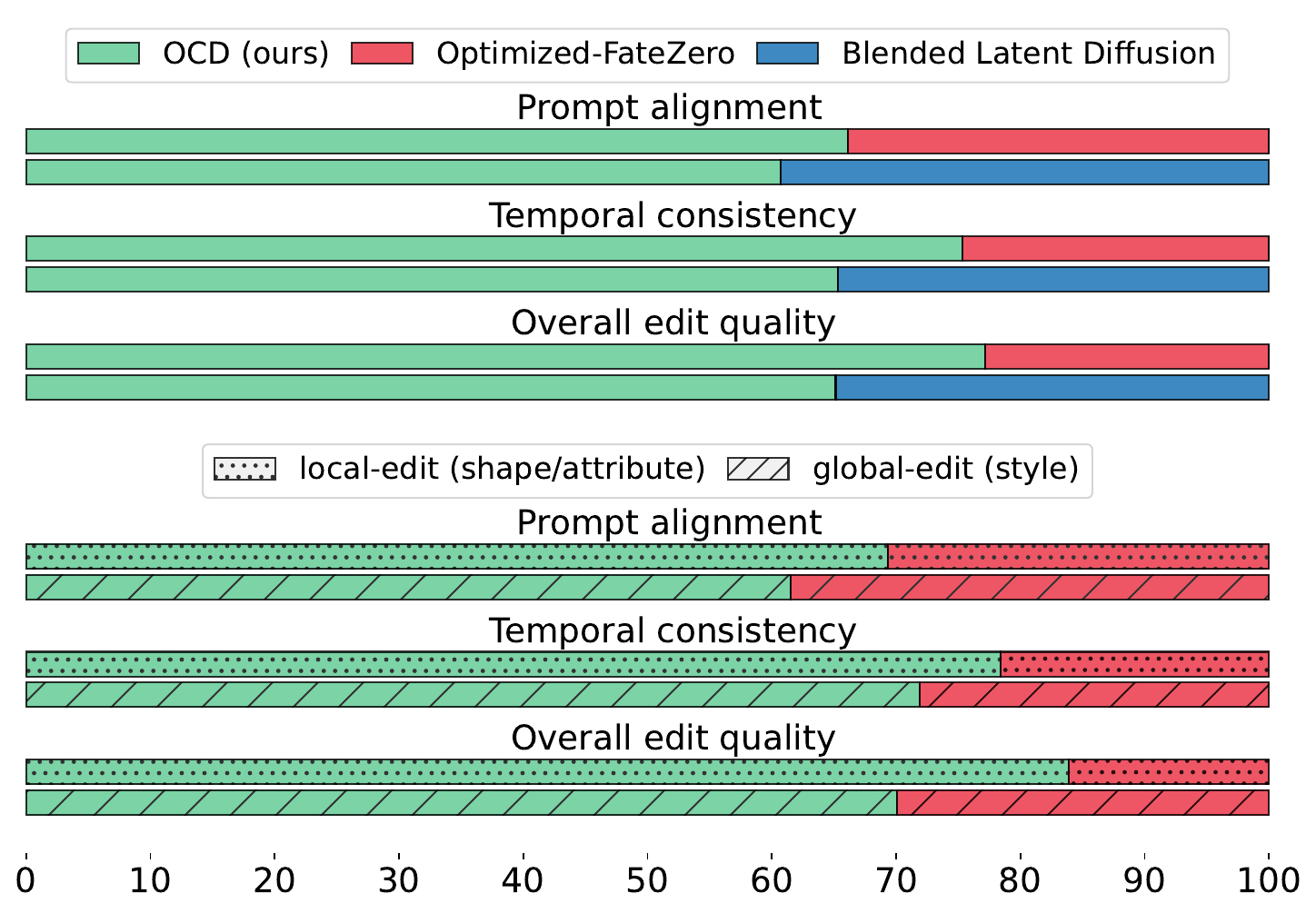}
\caption{
\textbf{User study:} (Top) Preferences for Object-Centric Diffusion (ours) w.r.t.~Optimized-FateZero or Blended Latent Diffusion \cite{avrahami2023blended} on FateZero benchmark (Bottom) Preferences for local vs.~global edits. In both cases, OCD is better preferred.}
\label{fig:user_study}
\end{figure}